\documentclass{article}

    \PassOptionsToPackage{numbers, compress}{natbib}

\usepackage[final]{neurips_2020}

\usepackage[utf8]{inputenc} 
\usepackage[T1]{fontenc}    
\usepackage[pagebackref=true,breaklinks=true,letterpaper=true,colorlinks,bookmarks=false]{hyperref}       
\usepackage{url}            
\usepackage{booktabs}       
\usepackage{amsfonts}       
\usepackage{nicefrac}       
\usepackage{microtype}      

\usepackage{graphicx}
\usepackage{natbib}
\usepackage{amsmath}
\usepackage{amssymb}
\usepackage{amsthm}
\newcommand{\etal}{\textit{et al.}}
\usepackage{algorithm}
\usepackage{algorithmic}

\newcommand{\norm}[1]{\left\lVert#1\right\rVert}

\newcounter{myequation}
\makeatletter
\@addtoreset{equation}{myequation}
\makeatother

\newcounter{myalgorithm}
\makeatletter
\@addtoreset{algorithm}{myalgorithm}
\makeatother

\newcounter{mytable}
\makeatletter
\@addtoreset{table}{mytable}
\makeatother

\newcounter{mysection}
\makeatletter
\@addtoreset{section}{mysection}
\makeatother

\newcounter{myfigure}
\makeatletter
\@addtoreset{figure}{myfigure}
\makeatother
\title{Guided Adversarial Attack for Evaluating and Enhancing Adversarial Defenses}

\author{Gaurang Sriramanan\thanks{Equal contribution. \newline Correspondence to: Gaurang Sriramanan <gaurangs@iisc.ac.in>,~Sravanti Addepalli  <sravantia@iisc.ac.in>}~, ~Sravanti Addepalli\footnotemark[1]~, ~Arya Baburaj, ~R.Venkatesh Babu\\
Video Analytics Lab, Department of Computational and Data Sciences\\ Indian Institute of Science, Bangalore, India\\}

\begin{document}

\maketitle

\begin{abstract}
  Advances in the development of adversarial attacks have been fundamental to the progress of adversarial defense research. Efficient and effective attacks are crucial for reliable evaluation of defenses, and also for developing robust models. Adversarial attacks are often generated by maximizing standard losses such as the cross-entropy loss or maximum-margin loss within a constraint set using Projected Gradient Descent (PGD). In this work, we introduce a relaxation term to the standard loss, that finds more suitable gradient-directions, increases attack efficacy and leads to more efficient adversarial training. We propose \emph{Guided Adversarial Margin Attack} (GAMA), which utilizes function mapping of the clean image to guide the generation of adversaries, thereby resulting in stronger attacks. We evaluate our attack against multiple defenses and show improved performance when compared to existing attacks. Further, we propose \emph{Guided Adversarial Training} (GAT), which achieves state-of-the-art performance amongst single-step defenses by utilizing the proposed relaxation term for both attack generation and training.
\end{abstract}

\section{Introduction}

The remarkable success of Deep Learning algorithms has led to a surge in their adoption in a multitude of applications which influence our lives in numerous ways. This makes it imperative to understand their failure modes and develop reliable risk mitigation strategies. One of the biggest known threats to systems that deploy Deep Networks is their vulnerability to crafted imperceptible noise known as adversarial attacks, as demonstrated by Szegedy \etal  \cite{intriguing-iclr-2014} in 2014. This finding has spurred immense interest towards identifying methods to improve the robustness of deep neural networks against adversarial attacks. While initial attempts of improving robustness against adversarial attacks used just single-step adversaries for training \cite{goodfellow2014explaining}, they were later shown to be ineffective against strong multi-step attacks by Kurakin \etal \cite{kurakin2016adversarial}. Some of the defenses introduced randomised or non-differentiable components, either in the pre-processing stage or in the network architecture, so as to minimise the effectiveness of generated gradients. However, many such defenses \cite{buckman2018thermometer, xie2018mitigating, song2018pixeldefend, guo2018countering} were later broken by Athalye \etal \cite{athalye2018obfuscated} using smooth approximations of the function during the backward pass or by computing reliable gradients using expectation over the randomized components. This game of building defenses against existing attacks, and developing attacks against the proposed defenses has been crucial for the progress in this field. Lately, the community has also recognized that the true testimony of a developed defense is to evaluate it against adaptive attacks which are constructed specifically to compromise the defense at hand \cite{carlini2019evaluating}. 

Multi-step adversarial training is one of the best known methods of achieving robustness to adversarial attacks today \cite{madry-iclr-2018,zhang2019theoretically}. This training regime attempts to solve the minimax optimization problem of firstly generating strong adversarial samples by maximizing a loss, and subsequently training the model to minimize loss on these adversarial samples. The effectiveness of the defense thus developed depends on the strength of the attack used for training. Therefore, development of stronger attacks is important for both evaluating existing defenses, and also for constructing adversarial samples during adversarial training. Indeed, the study of building robust adversarial defenses and strong adversarial attacks are closely coupled with each other today.

Adversarial attacks are constructed by maximizing standard losses such as cross-entropy loss or maximum-margin loss within a constrained set, as defined by the threat model. Due to the non-convex nature of the loss function, maximization of such a loss may not effectively find the path towards the class whose decision boundary is closest to the data point. 

In this work, we aid the optimization process by utilizing the knowledge embedded in probability values corresponding to non-maximal classes to guide the generation of adversaries. Motivated by graduated optimization methods, we improve the optimization process by introducing an $\ell_2$ relaxation term initially, and reducing the weight of this term gradually over the course of optimization, thereby making it equivalent to the primary objective towards the end. We demonstrate state-of-the-art results on multiple defenses and datasets using the proposed attack. We further analyse the impact of utilizing the proposed method to generate strong attacks for adversarial training. While use of the proposed attack for multi-step training shows only marginal improvement, we observe significant gains by using the proposed attack for single-step adversarial training. Single-step methods rely heavily on the initial gradient direction, and hence the proposed attack shows significant improvement over existing methods. 

Our contributions in this work can be summarized as follows:
\begin{itemize}
    \item We propose \emph{Guided Adversarial Margin Attack} (GAMA), which achieves state-of-the-art performance across multiple defenses for a single attack and across multiple random restarts.
    \item We introduce a multi-targeted variant GAMA-MT, which achieves improved performance compared to methods that utilize multiple targeted attacks to improve attack strength \cite{gowal2019alternative}. 
    \item We demonstrate that Projected Gradient Descent based optimization (GAMA-PGD) leads to stronger attacks when a large number of steps ($100$) can be used, thereby making it suitable for defense evaluation; whereas, Frank-Wolfe based optimization (GAMA-FW) leads to stronger attacks when the number of steps used for attack are severely restricted ($10$), thereby making it useful for adversary generation during multi-step adversarial training.
    \item We propose \emph{Guided Adversarial Training} (GAT), which achieves state-of-the-art results amongst existing single-step adversarial defenses. We demonstrate that the proposed defense can scale to large network sizes and to large scale datasets such as ImageNet-$100$.

\end{itemize}

Our code and pre-trained models are available here: \url{https://github.com/val-iisc/GAMA-GAT}.

\section{Preliminaries}

\label{Prelim}

\textbf{Notation:} In this paper, we consider adversarial attacks in the setting of image classification using deep neural networks. We denote a sample image as $x\in \mathcal{X}$, and its corresponding label as $y \in \{1, \dots, N \}$, where  $\mathcal{X} $ indicates the sample space and $N$ denotes the number of classes. Let $f_{\theta} :\mathcal{X} \rightarrow [0,1]^N$ represent the deep neural network with parameters ${\theta}$, that maps an input image $x$ to its softmax output $f_{\theta}(x) = \left( f_{\theta}^1(x), \dots , f_{\theta}^N(x)\right) \in [0,1]^N$. Further, let $C_{\theta}(x)$ represent the argmax over the softmax output. Thus, the network is said to successfully classify an image when $C_{\theta}(x) = y$. The cross-entropy loss for a data sample, $(x_i,y_i)$ is denoted by $\ell_{CE}(f_\theta(x_i),y_i)$. We denote an adversarially modified counterpart of a clean image $x$ as $\widetilde{x}$. 

\textbf{Adversarial Threat Model:} The goal of an adversary is to alter the clean input image $x$ such that the attacked image $\widetilde{x}$ is perceptually similar to $x$, but causes the network to misclassify. Diverse operational frameworks have been developed to quantify perceptual similarity, and adversarial attacks corresponding to these constraints have been studied extensively. We primarily consider the standard setting of worst-case adversarial attacks, subject to $\ell_p$-norm constraints. More precisely, we consider adversarial threats bound in $\ell_{\infty}$ norm: 
 $ \widetilde{x} \in \{x': \| {x' - x}\|_{\infty} \le \varepsilon \}$.

While evaluating the proposed defense, we consider that the adversary has full access to the model architecture and parameters, since we consider the setting of worst-case robustness. Further, we assume that the adversary is cognizant of the defense techniques utilised during training or evaluation.

\section{Related Works}
\label{related_works}
\subsection{Adversarial Attacks}
A panoply of methods have been developed to craft adversarial perturbations under different sets of constraints. One of the earliest attacks specific to $\ell_{\infty}$ constrained adversaries was the Fast Gradient Sign Method (FGSM), introduced by Goodfellow \etal \cite{goodfellow2014explaining}. In this method, adversaries are generated using a single-step first-order approximation of the cross-entropy loss by performing simple gradient ascent. Kurakin \etal  \cite{kurakin2016physical} introduced a significantly stronger, multi-step variant of this attack called Iterative FGSM (I-FGSM), where gradient ascent is iteratively performed with a small step-size, followed by re-projection to the constraint set. Madry \etal \cite{madry-iclr-2018} developed a variant of this attack, which involves the addition of initial random noise to the clean image, and is commonly referred to as Projected Gradient Descent (PGD) attack. 

Carlini and Wagner \cite{carlini2017towards} explored the use of different surrogate loss functions and optimization methods to craft adversarial samples with high fidelity and small distortion with respect to the original image. The authors introduce the use of maximum margin loss for generation of stronger attacks, as opposed to the commonly used cross-entropy loss. Our proposed attack introduces a relaxation term in addition to the maximum margin loss in order to find more reliable gradient directions. 

The Fast Adaptive Boundary (FAB) attack, introduced by Croce and Hein \cite{croce2019minimally} produces minimally distorted adversarial perturbations with respect to different norm constraints, using a linearisation of the network followed by gradient steps which have a bias towards the original sample. While the FAB attack is often stronger than the standard PGD attack, it is computationally more intensive for the same number of iterations. Gowal \etal \cite{gowal2019alternative} introduced the Multi-Targeted attack, which cycles over all target classes, maximising the difference of logits corresponding to the true class and the target class. While this attack finds significantly stronger adversaries compared to PGD attack, it relies on cycling over multiple target classes, and hence requires a large computational budget to be effective. More recently, Croce and Hein \cite{croce2020reliable} proposed AutoPGD, which is an automatised variant of the PGD attack, that uses a step-learning rate schedule adaptively based on the past progression of the optimization. They further introduce a new loss function, the Difference of Logits Ratio (DLR), which is a scale invariant version of the maximum margin loss on logits, and outperforms the $\ell_\infty$ based Carlini and Wagner (C\&W) attack \cite{carlini2017towards}. Additionally, they proposed AutoAttack, an ensemble of AutoPGD with the cross-entropy loss and the DLR loss, the FAB attack and Square attack \cite{andriushchenko2019square}, a score-based black-box attack which performs zeroth-order optimization.
\subsection{Defenses Against Adversarial Attacks}
With the exception of a few defenses \cite{cohen2019certified,addepalli2020bpfc}, most methods used to produce robust networks include some form of adversarial training, wherein training data samples are augmented with adversarial samples during training. Early works proposed training on FGSM \cite{goodfellow2014explaining}, or Randomised FGSM (R-FGSM) \cite{tramer2017ensemble} adversaries to produce robust networks. However, these models were still overwhelmingly susceptible to multi-step attacks \cite{kurakin2016adversarial} due to the Gradient Masking effect \cite{papernot2017practical}. Madry \etal \cite{madry-iclr-2018} proposed a min-max formulation for training adversarially robust models using empirical risk minimisation. It was identified that strong, multi-step adversaries such as Projected Gradient Descent (PGD), were required to sufficiently approximate the inner maximization step, so that the subsequent adversarial training yields robust models. Following this, Zhang \etal \cite{zhang2019theoretically} presented a tight upper bound on the gap between natural and robust error, in order to quantify the trade-off between accuracy and robustness. Using the theory of classification calibrated losses, they develop TRADES, a multi-step gradient-based technique. However, methods such as TRADES and PGD-Training are computationally intensive, as they inherently depend upon the generation of strong adversaries through iterative attacks.

Consequently, efforts were made to develop techniques that accelerated adversarial training. Shafahi \etal  \cite{shafahi2019adversarial} proposed a variant of PGD-training, known as Adversarial Training for Free (ATF), where the gradients accumulated in each step are used to simultaneously update the adversarial sample as well as network parameters, enabling the generation of strong adversaries during training, without additional computational overheads. 

In order to mitigate gradient masking as seen in prior works that used single-step attacks for adversarial training, Vivek \etal \cite{baburaj2019regularizer} proposed the use of the R-MGM regularizer. The authors minimize the squared $\ell_2$ norm of the difference between logits corresponding to FGSM and R-FGSM adversaries to train adversarially robust models. In contrast to this, we introduce a regularizer to minimize the squared $\ell_2$ distance between the softmax outputs of clean and adversarial images, thereby improving the computational efficiency. Secondly, the adversary generation process uses the proposed Guided Adversarial Attack, thereby resulting in the use of a significantly stronger attack during training. 

Contrary to prior wisdom,  Wong \etal \cite{wong2020fast} (FBF), found the surprising result that R-FGSM training could indeed be successfully utilised to produce robust models. It was shown that R-FGSM adversarial training could be made effective with the use of small-step sizes for generation of adversaries, in combination with other techniques such as early-stopping and cyclic learning rates. With these techniques, they obtain better performance when compared to Adversarial Training for Free, with further reduction in computational requirements. While our proposed defense is also based on adversarial training with single-step adversaries, our choice of the loss function enables generation of stronger adversaries, thereby resulting in models that are significantly more robust. Further, we note that the acceleration techniques used in \cite{wong2020fast} can be utilized for our method as well. 

\section{Proposed Method}

\begin{figure}
\centering
        \includegraphics[width=\linewidth]{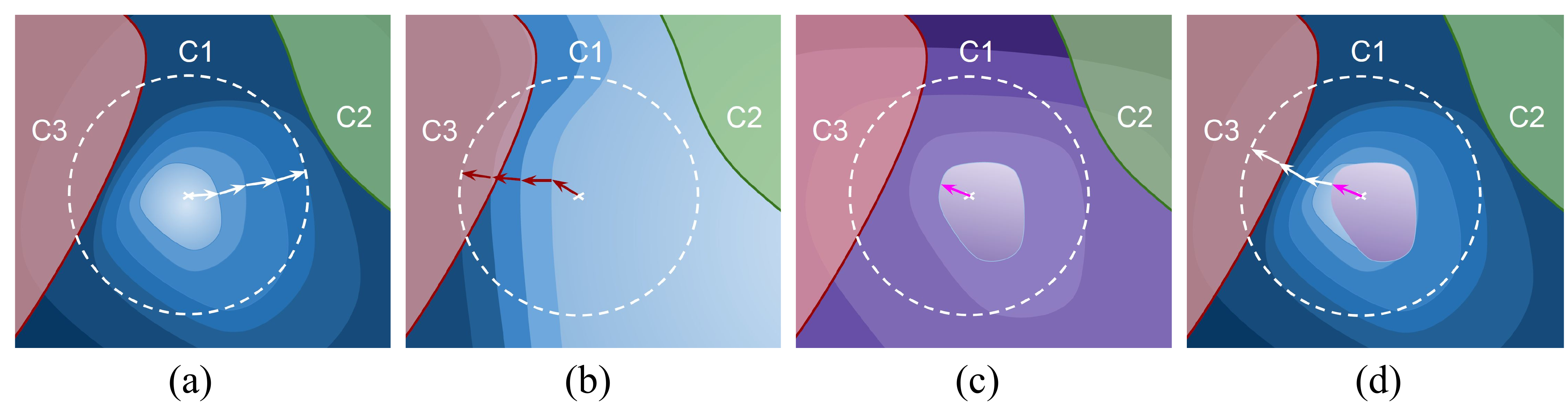}
        \vspace{-0.3cm}
        \caption{Schematic diagram of loss contours (a) Untargeted loss (b) Targeted loss w.r.t. class C3 (c) Guided loss for initial optimization (d) Path of adversary using GAMA}
        \label{fig:loss_contours}
        \vspace{-0.3cm}
\end{figure}

\label{proposed_method}

\subsection{Impact of Initial Optimization Trajectory on Attack Efficacy}
One of the most effective attacks known till date is the Projected Gradient Descent (PGD) attack \cite{madry-iclr-2018}, which starts with a random initialization and moves along the gradient direction to maximize cross entropy loss. Each iteration of PGD takes a step of a fixed size in the direction of sign of the gradient, after which the generated perturbation is projected back to the epsilon ball. 
Owing to the non-convex nature of the loss function, the initial gradient direction that maximizes cross-entropy loss may not lead to the optimal solution. This could lead to the given data sample being correctly classified, even if adversaries exist within an epsilon radius. This is shown in the schematic diagram of loss contours in Fig.\ref{fig:loss_contours}(a), where the adversary moves towards class C2 based on the initial gradient direction, and fails to find the adversary that belongs to class C3.

This is partly mitigated by the addition of initial random noise, which increases the chance of the adversary moving towards different directions. However, this gain can be seen only when the attack is run for multiple random restarts, thereby increasing the computational budget required for finding an adversarial perturbation. 
Another existing approach that gives a better initial direction to the adversaries is the replacement of the standard untargeted attack with a combination of multiple targeted attacks \cite{gowal2019alternative}. This diversifies the initial direction of adversaries over multiple random restarts, thereby resulting in a stronger attack. This can be seen in Fig.\ref{fig:loss_contours}(b), where the adversary is found by minimizing a targeted loss corresponding to the class C3, which has the closest decision boundary to the given sample. While this is a generic approach which can be used to strengthen any attack (including GAMA), it does not scale efficiently as the number of target  classes increase. 

In this paper, we propose to utilize supervision from the function mapping of clean samples in order to identify the initial direction that would lead to a stronger attack (Fig.\ref{fig:loss_contours}(c)). The proposed attack achieves an effect similar to the multi-targeted attack without having to explicitly minimize the loss corresponding to each class individually (Fig.\ref{fig:loss_contours}(d)). This leads to more reliable results in a single, or very few restarts of the attack, thereby improving the scalability of the attack to datasets with larger number of classes. 

\begin{figure}

\centering
        \includegraphics[width=\linewidth]{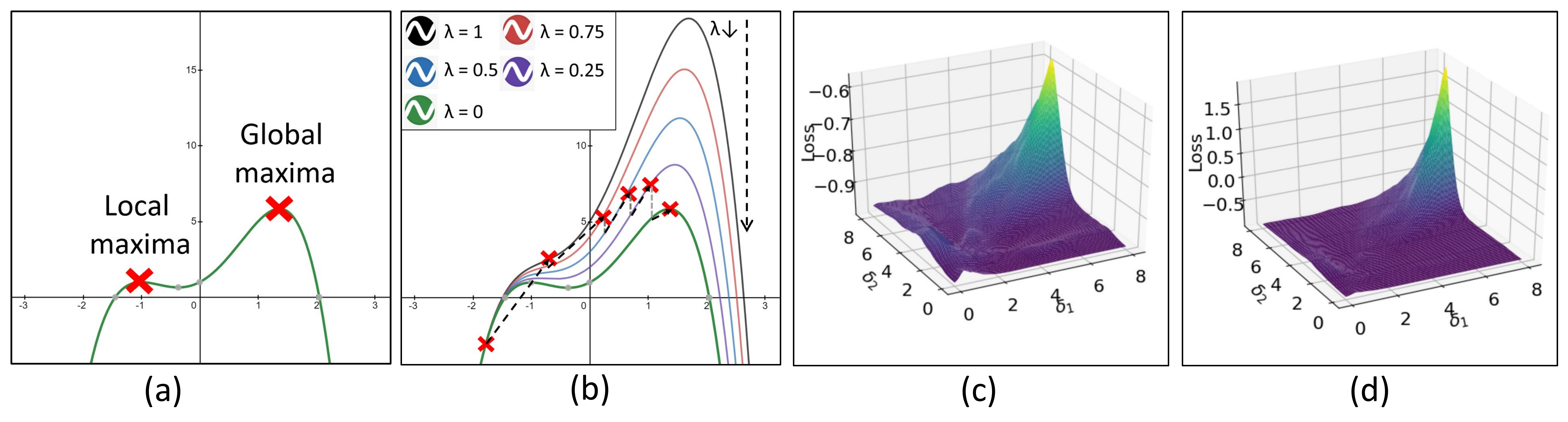}
        \vspace{-0.3cm}
        \caption{Addition of decaying $\ell_2$ relaxation term for function smoothing and improved optimization (a) 1-D example showing maximization of the non-concave function, $1-x(x-2)(x+1)(x+1)$ (b) The smoothed function after addition of $\ell_2$ relaxation term is $1-x(x-2)(x+1)(x+1)+\lambda(x+2)^2$. $\lambda$ is reduced from $1$ to $0$ over iterations. Optimization trajectory is shown using black dotted lines and red cross marks. (c, d) Plot of the loss surface of an FGSM trained model on perturbed images of the form $x^* = x+\delta_1 g+\delta_2 g^\perp$, obtained by varying $\delta_1$ and $\delta_2$. Here $g$ is the sign of the gradient direction of the loss with respect to the clean image ($x$) and $g^\perp$ is a direction orthogonal to $g$. Loss functions used are: (c) Maximum-margin loss, and (d) GAMA loss as shown in Eq.\ref{Attack_loss}, with $\lambda$ set to 25. Addition of the relaxation term helps in smoothing the loss surface and suppressing gradient masking.}
        \label{fig:graduated_opt_fig}
        \vspace{-0.3cm}
\end{figure}

\subsection{Guided Adversarial Margin Attack}
\label{subsec:GAMA}

Due to the inherent difficulty observed in the optimization of non-convex functions, several heuristic methods such as Graduated Optimization have been deployed to obtain solutions that sufficiently approximate global optima. To optimize a non-convex function, Graduated methods attempt to construct a family of smooth function approximations which are more amenable to standard optimization techniques. These function approximations are progressively refined in order to recover the original function toward the end of optimization. Hazan \etal \cite{hazan2016graduated} proposed to utilise projected gradient descent with a noisy gradient oracle, to optimize graduated function approximations obtained by local averaging over progressively shrinking norm balls. The authors characterise a family of functions for which their algorithm recovers approximate solutions of the global optima. 

Along similar lines, we seek to introduce a relaxation term to obtain a series of smooth approximations of the primary objective function that is used to craft adversarial perturbations. We illustrate a simplified $1$-dimensional example in Fig.\ref{fig:graduated_opt_fig}(a,b) to highlight the efficacy of graduated optimization through function smoothing. The loss function that is maximized for the generation of the proposed Guided Adversarial Margin Attack (GAMA) is as follows: 
\begin{equation}
\label{Attack_loss}
    L =  - f_{\theta}^{y} (\widetilde{x}) + \max_{j \neq y}   f_{\theta}^{j} (\widetilde{x})  +  \lambda \cdot   || \boldsymbol{f_{\theta}} (\widetilde{x}) -  \boldsymbol{f_{\theta}} (x)   ||_2^2
\end{equation}

The first two terms in the loss correspond to the maximum margin loss in probability space, which is the difference between the probability score of the true class $f_{\theta}^{y} (x)$, and the probability score of the second most confident class $j \neq y$. The standard formulation of PGD attack maximizes cross-entropy loss for the generation of attacks. We use maximum-margin loss here, as it is known to generate stronger attacks when compared to cross-entropy loss \cite{carlini2017towards,gowal2019alternative}. In addition to this, we introduce a relaxation term corresponding to the squared $\ell_2$ distance between the probability vectors $f_{\theta}$ of the clean image $x$ and the perturbed image $\widetilde{x}$. This term is weighted by a factor $\lambda$ as shown in Eq.\ref{Attack_loss}. Similar to graduated optimization, this weighting factor is linearly decayed to $0$ over iterations, so that this term only aids in the optimization process, and does not disturb the optimal solution of the true maximum-margin objective. As shown in Fig.\ref{fig:graduated_opt_fig}(c,d), the $\ell_2$ relaxation term indeed leads to a smoother loss surface in an FGSM trained model. 

The gradients of this $\ell_2$ relaxation term are a weighted combination of the gradients of each of the class confidence scores of the perturbed image. Each term is weighted by the difference in corresponding class confidence scores of the perturbed image and clean image. Therefore, a direction corresponding to the gradient of a given class confidence score is given higher importance if it has already deviated by a large amount from the initial class confidence of the clean image. Thus, the weighting of the current gradient direction considers the cumulative effect of the previous steps, bringing about an advantageous effect similar to that of momentum. This helps direct the initial perturbation more strongly towards the class which maximizes the corresponding class confidence, while also making the optimization more robust to spurious random deviations due to local gradients. 

The algorithm for the proposed attack is presented in Algorithm-\ref{alg:GAMA}. The attack is initialized using random Bernoulli noise of magnitude $\varepsilon$. This provides a better initialization when compared to Uniform or Gaussian noise, as the resultant image would be farther away from the clean image in this case when compared to other methods, resulting in more reliable gradients initially. Secondly, the space of all sign gradient directions is represented completely by the vertices of the $\ell_\infty$ hypercube of a fixed radius around the clean image, which is uniformly explored using Bernoulli noise initialization. The attack is generated using an iterative process that runs over $T$ iterations, where the current step is denoted by $t$. At each step, the loss in Eq.\ref{Attack_loss} is maximized to find the optimal $\widetilde{x}$ for the given iteration. The weighting factor $\lambda$ of the $\ell_2$ term in the loss function is linearly decayed to $0$ over $\tau$ steps. 

We propose two variants of the Guided Adversarial Margin Attack, GAMA-PGD and GAMA-FW. GAMA-PGD uses Projected Gradient Descent for optimization, while GAMA-FW uses the Frank-Wolfe \cite{frank1956algorithm} algorithm, also known as Conditional Gradient Descent. In PGD, the constrained optimization problem is solved by first posing the same as an unconstrained optimization problem, and further projecting the solution onto the constraint set. Gradient ascent is performed by computing the sign of the gradient, and taking step of size $\eta$, after which the perturbation is clamped between $-\varepsilon$ and $\varepsilon$, to project to the $\ell_\infty$ ball. On the other hand, the Frank-Wolfe algorithm finds the optimal solution in the constraint set by iteratively updating the current solution as a convex combination of the present perturbation and the point within the constraint set that maximises the inner-product with the gradient. For the setting of $\ell_{\infty}$ constraints, this point which maximises the inner-product is simply given by epsilon times the sign of the current gradient. Since the constraint set is convex, this process ensures that the generated solution lies within the set, and hence does not require a re-projection to the same. This process results in a faster convergence, thereby resulting in stronger attacks when the budget for the number of iterations is small. This makes GAMA-FW particularly useful in the setting of adversarial training, where there is a fixed budget on the number of steps used for attack generation. Finally the image is clamped to be in the range $[0,1]$. We use an initial step size of $\eta$ for GAMA-PGD and $\gamma$ for GAMA-FW, and decay this by a factor of $d$ at intermediate steps. 
\begin{algorithm}[tb]
   \caption{Guided Adversarial Margin Attack }
   \label{alg:GAMA}
   
\begin{algorithmic}[1]
   \STATE {\bfseries Input:} Network $f_{\theta}$ with parameters $\theta$, Input image $x$ with label $y$, Attack Size $\varepsilon$,   Step-size $\eta$ or Convex parameter $\gamma$, Initial Weighting factor $\lambda_0$, Total Steps $T$, Relaxation Steps $\tau$, Learning Rate Schedule $S = \{t_1, \dots, t_k \}$, Learning Rate Drop Factor $d$
   \STATE {\bfseries Output:} Adversarial Image $\widetilde{x}_T$

\STATE $\delta = Bern(-\varepsilon,\varepsilon)$   \small{       ~~~~~~~~~~~~~~~~~~~~~~~~~~// Initialise Perturbation with Bernoulli Noise}
\STATE $\widetilde{x}_0 = x_0 = x + \delta$~, $\lambda = \lambda_0$
\FOR{$t=0$ {\bfseries to} $T-1$}

    \STATE $L =   \max_{j \neq y} \{f_{\theta}^{j} (\widetilde{x}_t)\} - f_{\theta}^{y} (\widetilde{x}_t) +  \lambda \cdot   || f_{\theta} (\widetilde{x}_t) -  f_{\theta} (x_0)   ||_2^2 $
    \STATE $\lambda = \max (\lambda - \lambda_0/\tau,0 $ )

    \IF{mode == PGD}
        \STATE $\delta = \delta +  \eta \cdot sign(\nabla_{\delta} L)$ 
        \STATE $\delta = Clamp(\delta,-\varepsilon,\varepsilon)$ \small{     ~~~~~~~~~~~~~~// Project Back To Constraint Set}
    \ELSIF{mode == FW}
        \STATE $\delta = (1-\gamma) \cdot \delta +  \gamma \cdot \varepsilon\cdot sign(\nabla_{\delta} L)$
    \ENDIF
    
    \STATE $\delta = Clamp(x+\delta,0,1) - x$    
     \STATE $\widetilde{x}_{t+1} = x + \delta $

    \IF{$t \in S$} 
            \STATE  $\eta = \eta /d$~, $\gamma = \gamma/d$
    \ENDIF
    
\ENDFOR

\end{algorithmic}
\end{algorithm}

\subsection{Guided Adversarial Training}
\label{main_sec:GAT}
In this section, we discuss details on the proposed defense GAT, which utilizes single-step adversaries generated using the proposed Guided Adversarial attack for training. As discussed in Section-\ref{subsec:GAMA}, the $\ell_2$ term between the probability vectors of clean and adversarial samples in Eq.\ref{Attack_loss} provides reliable gradients for the optimization, thereby yielding stronger attacks in a single run. The effectiveness and efficiency of the proposed attack make it suitable for use in adversarial training, to generate more robust defenses. This attack is notably more useful for training single-step defenses, where reliance on the initial direction is significantly higher when compared to multi-step attacks.

Initially, Bernoulli noise of magnitude $\alpha$ is added to the input image in order to overcome any possible gradient masking effect in the vicinity of the data sample. Next, an attack is generated by maximizing loss using single step optimization. We use the minimax formulation proposed by Madry \etal \cite{madry-iclr-2018} for adversarial training, where the maximization of a given loss is used for the generation of attacks, and minimization of the same loss on the generated adversaries leads to improved robustness. In order to use the same loss for both attack generation and training, we use cross-entropy loss instead of the maximum-margin loss in Eq.\ref{Attack_loss}. This improves the training process, as cross-entropy loss is known to be a better objective for training when compared to maximum-margin loss.
The generated perturbation is then projected onto the $\varepsilon$-ball. We introduce diversity in the generated adversaries by setting $\lambda$ to $0$ in alternate iterations, only for the attack generation. These adversarial samples ($\widetilde{x_i}$) along with the clean samples ($x_i$) are used for adversarial training. The algorithm of the proposed single-step defense GAT is presented in detail in Algorithm-\ref{alg:GAT} of the Supplementary section.

Single-step adversarial training methods commonly suffer from gradient masking, which prevents the generation of strong adversaries, thereby leading to weaker defenses. The proposed training regime caters to the dual objective of minimizing loss on adversarial samples, while also explicitly enforcing function smoothing in the vicinity of each data sample (Details in Section-\ref{sec:local_prop} of the Supplementary section). The latter outcome strengthens the credibility of the linearity assumption used during generation of single-step adversaries, thereby improving the efficacy of the same. This coupled with the use of stronger adversaries generated using GAMA enables GAT to achieve state-of-the-art robustness among the single-step training methods.

\vspace{-0.05cm}
\section{Experiments and Analysis}
\label{experiments}
In this section, we present details related to the experiments conducted to validate our proposed approach. We first present the experimental results of the proposed attack GAMA, followed by details on evaluation of the proposed defense GAT. The primary dataset used for all our evaluations is CIFAR-$10$ \cite{krizhevsky2009learning}. We also show results on MNIST \cite{lecun1998mnist} and ImageNet \cite{imagenet_cvpr09} for the proposed attack GAMA in the main paper and for the proposed defense GAT in Section-\ref{sec:eval_defense} of the Supplementary. We use the constraint set given by the $\ell_{\infty}$ ball of radius $8/255$, $8/255$ and $0.3$ for the CIFAR-$10$, ImageNet and MNIST datasets respectively. The implementation details of the proposed defense and attack are presented in Sections-\ref{sec:GAT} and \ref{sec:impl_gama} of the Supplementary. 
\begin{table}[t]
\caption{\textbf{Attacks (CIFAR-10)}: Accuracy ($\%$) of various defenses (rows) against adversaries generated using different 100-step attacks (columns) under the $\ell_\infty$ bound with $\varepsilon=8/255$. Architecture of each defense is described in the column "Model". WideResNet is denoted by W (W-28-10 represents WideResNet-28-10), ResNet-18 is denoted by RN18, Pre-Act-ResNet-18 is denoted by PA-RN18.   $^\ddagger$Additional data used for training, $^\dagger\varepsilon=0.031$, $^\ast$Defenses trained using single-step adversaries}
\setlength\tabcolsep{3pt}
\resizebox{1.0\linewidth}{!}{
\label{table:attacks_main}
\begin{tabular}{l|c|cccccc|cccc|cc}
\toprule
\multicolumn{1}{l|}{}  &     & \multicolumn{6}{c|}{\textbf{Single run of the attack}}                            & \multicolumn{4}{c}{\textbf{5 random restarts}}&\multicolumn{2}{|c}{\textbf{Top 5 targets}}\\
\multicolumn{1}{l|}{}   &  \textbf{Model}  & \small{PGD} & \small{APGD} & \small{APGD} & \small{FAB}   & \small{GAMA} & \small{GAMA} & \small{APGD} & \small{FAB}   & \small{GAMA} & \small{GAMA} &  \small{MT} & \small{GAMA}\\
\multicolumn{1}{l|}{}    &  & \small{100} & \small{CE} & \small{DLR} &   & \small{PGD} & \small{FW} & \small{DLR} &     & \small{PGD} & \small{FW} & & \small{PGD-MT}\\
\midrule
Carmon \etal \cite{carmon2019unlabeled}$^\ddagger$  & W-28-10 & 61.86   & 61.81   & 60.85    & 60.88 & \textbf{59.81}  & 59.83          & 60.64    & 60.62         & \textbf{59.65}  & 59.71   & 59.86   &  \textbf{59.56}  \\
Sehwag \etal \cite{sehwag2020pruning}$^\ddagger$ &  W-28-10 & 59.93   & 59.61   & 58.39    & 58.29 & 57.51           & \textbf{57.50}  & 58.26    & 58.06         & \textbf{57.37}  & 57.38  & 57.48   &  \textbf{57.20}   \\
Wang \etal \cite{Wang2020Improving}$^\ddagger$  &  RN18  & 52.87   & 52.38   & 49.70     & 48.50  & \textbf{48.12}  & 48.17          & 49.37    & 48.33         & \textbf{47.92}  & 47.97 & 47.76  &   \textbf{47.58}    \\
Wang \etal \cite{Wang2020Improving}$^\ddagger$   & W-28-10  & 62.63   & 61.76   & 58.98    & 57.53 & 57.19           & \textbf{57.14} & 58.56    & 57.29         & \textbf{56.84}  & 56.92  & 56.80   &  \textbf{56.54}   \\
Hendrycks \etal \cite{hendrycks2019using}$^\ddagger$ & W-28-10 & 57.58   & 57.20    & 57.25    & 55.55 & 55.24           & \textbf{55.19} & 56.96    & 55.40          & 55.11           & \textbf{55.08} & 55.06 & \textbf{54.92}\\
Rice \etal \cite{rice2020overfitting}  &   W-34-20  & 57.25   & 56.93   & 55.99    & 54.34 & \textbf{53.77}  & 53.88          & 55.70     & 54.19         & \textbf{53.64}  & 53.68    & 53.59   &  \textbf{53.45} \\
Zhang \etal \cite{zhang2019theoretically}$^\dagger$ & W-34-10 &  55.60  & 55.30    & 54.18    & 53.92 & \textbf{53.29}  & 53.38          & 54.04    & 53.82        & \textbf{53.17}  & 53.22  & 53.32    &  \textbf{53.09}   \\
Madry \etal \cite{madry-iclr-2018} \cite{madrylab}   & RN-50 & 53.49   & 51.78   & 53.03    & 50.67 & \textbf{50.04}  & 50.08          & 52.64    & 50.37         & \textbf{49.81}  & 49.92   & 49.76    &  \textbf{49.41} \\
Wong \etal \cite{wong2020fast}$^\ast$    & PA-RN18  & 46.42   & 45.96   & 46.95    & 44.51 & \textbf{43.85}  & 43.90           & 46.64    & 44.03        & \textbf{43.65}  & 43.69  & 43.65   &  \textbf{43.33}    \\
GAT (Ours)$^\ast$   &     W-34-10        & 55.10    & 54.73   & 53.08    & 51.28 & \textbf{50.76}  & 50.79          & 52.75    & 51.07        & \textbf{50.43}  & 50.48     & 50.45     &  \textbf{50.18}   \\
\midrule 
\end{tabular}}
\vspace{-0.3cm}
\end{table}

\begin{figure}
\centering
        \includegraphics[width=\linewidth]{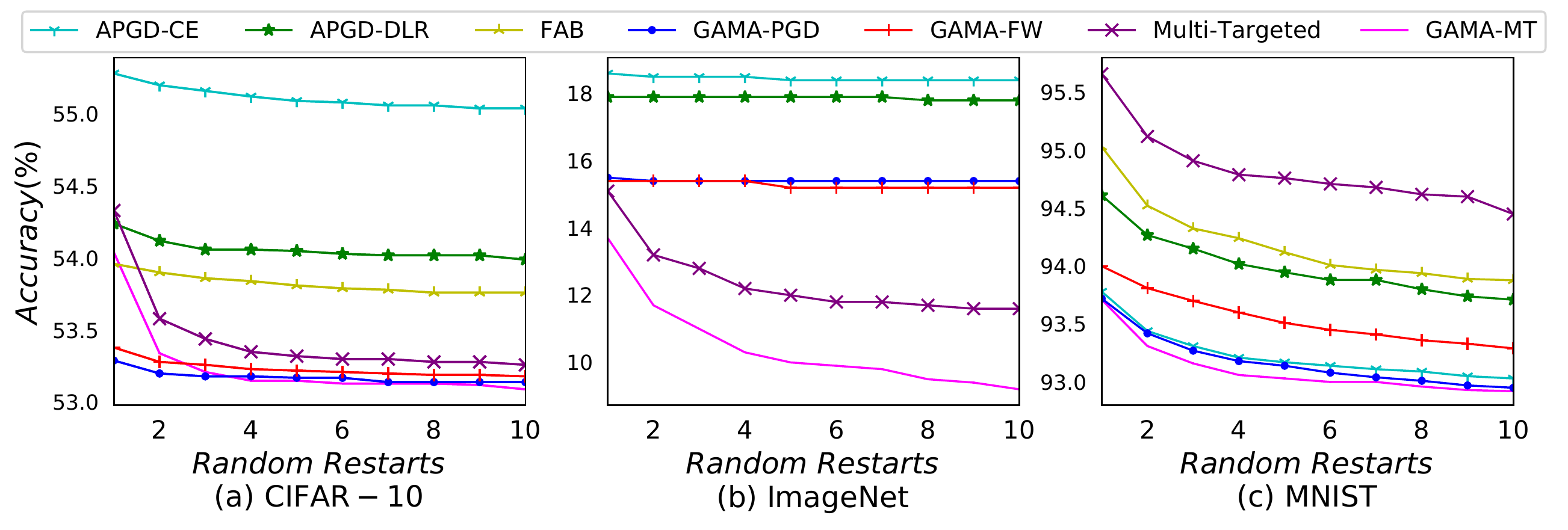}
        \vspace{-0.4cm}
        \caption{Accuracy ($\%$) of different attacks against multiple random restarts. Evaluations are performed on TRADES WideResNet-34 model \cite{zhang2019theoretically} for CIFAR-10, Madry \etal \cite{madrylab} ResNet-50 model for ImageNet (first 1000 samples), and TRADES SmallCNN \cite{zhang2019theoretically} model for MNIST.
        \label{fig:acc_vs_rr}}
        \vspace{-0.3cm}
\end{figure}

\subsection{Evaluation of the proposed attack (GAMA)}
The performance of various defenses against different attack methods on CIFAR-$10$ dataset is shown in Table-\ref{table:attacks_main}. We present results for both a single run of the attack (with a budget of 100 iterations), as well as the worst-case accuracy across 5 random restarts (with an effective budget of 5$\times$100 iterations). Notably, GAMA-PGD and GAMA-FW consistently outperform all other untargeted attacks across all defenses. Further, we remark that while the FAB attack stands as the runner-up method, it requires significantly more computation time, approximately $6$ times that of GAMA-PGD and GAMA-FW. 

The Multi-Targeted attack (MT) is performed by targeting the top 5 classes excluding the correct class. We present GAMA-MT, a multi-targeted version of the GAMA-PGD attack, where the maximum-margin loss is replaced by the margin loss targeted towards the top 5 classes excluding the true class. We note that the GAMA-MT attack is consistently the most effective attack across all defenses.

We further present evaluations on the TRADES WideResNet-34 model \cite{zhang2019theoretically}, PGD adversarially trained ResNet-50 model \cite{madry-iclr-2018} and TRADES SmallCNN model \cite{zhang2019theoretically} on the CIFAR-$10$, ImageNet (first 1000 samples) and MNIST datasets respectively against different attack methods in Fig.\ref{fig:acc_vs_rr}. We find that while GAMA-PGD and GAMA-FW continue to consistently achieve the strongest attacks, they are also less sensitive to the random initialisation, when compared to other attack methods for varying number of random restarts. Thus the proposed attacks offer a more reliable bound on the robustness of models, within a single restart or very few restarts. The proposed multi-targeted attack GAMA-MT outperforms all other attacks significantly on ImageNet, and is marginally better than GAMA-PGD for CIFAR-$10$ and MNIST.  

We evaluate the proposed attack on the TRADES leaderboard models \cite{zhang2019theoretically}. A multi-targeted version of our attack GAMA on the WideResNet-34 CIFAR-$10$ model achieved the top position in the leaderboard, with $53.01\%$ for a $100$-step attack with $20$ random restarts. On the SmallCNN MNIST model, we achieve an accuracy of $92.57\%$ for a $100$-step attack with $1000$ random restarts.

\textbf{Ablation Experiments:} We present evaluations on the TRADES WideResNet-34 model on the CIFAR-$10$ test set with several ablations of the proposed attack in Table-\ref{table:ablations} of the Supplementary section. We first observe that the maximum-margin loss is more effective when compared to the cross-entropy loss, for both $10$ and $100$ step attacks. Further, we observe that we obtain stronger adversaries while optimizing the margin loss between predicted probability scores, as compared to the corresponding logits. The weighting factor for the squared $\ell_2$ relaxation term is linearly decreased to $0$ for the $100$-step attack, while it is kept constant for the $10$-step attack. From the $100$-step evaluations, we observe that the graduated optimization indeed aids in finding stronger adversaries. Further, the addition of initial Bernoulli random noise aids in improving $100$-step adversaries. We also note that GAMA-FW achieves the strongest attack when the available budget on the number of steps for attack is relatively small. 

\begin{table}[t]
\caption{\textbf{Defenses (CIFAR-10)}: Accuracy ($\%$) of different models (rows) against various $\ell_\infty$ norm bound ($\varepsilon=8/255$, $^\dagger\varepsilon=0.031$) white-box attacks (columns). The first partition corresponds to single-step defenses, and the second has multi-step defenses. For the C\&W attack, the mean $\ell_2$ norm required to achieve high Fooling Rate (FR) is reported. Higher the $\ell_2$ norm, better is the robustness.}
\setlength\tabcolsep{7pt}
\resizebox{1.0\linewidth}{!}{
\label{table:cifar10_whitebox}
\begin{tabular}{l|c|c|c|c|ccc|c|c|c}
\toprule
& &\textbf{Clean~} & \multicolumn{6}{c|}{\textbf{Acc (\%) on $\ell_{\infty}$ attacks}} &&\multicolumn{1}{|c}{\textbf{C \& W}}\\ 
\cmidrule(l){4-9}
\textbf{Method} & \textbf{Model}   & \textbf{Acc (\%)} & \textbf{FGSM}  &\textbf{IFGSM}& \multicolumn{3}{c|}{\textbf{PGD (n-steps)}}&\textbf{GAMA}&\textbf{AA} & \textbf{Mean}        \\
                     &        &        &   & \textbf{7-step}    &\textbf{~~7~~}    & \textbf{~~20~~} & \textbf{~500~} &\textbf{PGD-100}& &\textbf{$\ell_2$} \\ \midrule

Normal     &    RN18    & 92.30 & 15.98                & 0.00  & 0.00  & 0.00  & 0.00 & 0.00 & 0.00     & 0.108                \\
FGSM-AT \cite{goodfellow2014explaining}    &   RN18    & 92.89 & \textbf{96.94}                & 0.82  & 0.38  & 0.00  & 0.00 & 0.00 & 0.00     & 0.078                \\
RFGSM-AT  \cite{tramer2017ensemble}   &   RN18   & 89.24 & 49.94                & 42.52 & 41.02 & 35.02 & 34.17 & 33.87 & 33.16 & 0.634                \\
ATF    \cite{shafahi2019adversarial}     &   RN18    & 71.77 & 46.67                & 45.06 & 44.96 & 43.53 & 43.52 &  40.34 &  40.22   & 0.669                \\
FBF  \cite{wong2020fast}     &    RN18     & 82.83 & 54.09                & 50.28 & 49.66 & 46.41 & 46.03 & 43.85 &  43.12    & 0.685                \\
R-MGM  \cite{baburaj2019regularizer}    &   RN18     & 82.29 & 55.04                & 50.87 & 50.03 & 46.23 & 45.79 & 44.06 & 43.72 & 0.745                \\
GAT (\textbf{Ours})     &  RN18      & 80.49 & 57.37                & 55.32 & 54.99 & 53.13 & 53.08 & 47.76 & 47.30 & \textbf{0.762}                \\
FBF \cite{wong2020fast}   &  WRN34   & 82.05 &           53.79           &   49.20    & 49.51 & 46.35 & 45.94 &  43.13 & 43.14    &        0.628              \\
GAT (\textbf{Ours})  & WRN34   & 85.17 & 61.93                & \textbf{58.68} & \textbf{57.25} & \textbf{55.34} & \textbf{55.10} & \textbf{50.76} & \textbf{50.27} &   0.724                   \\
\midrule
PGD-AT  \cite{madry-iclr-2018}    &   RN18    & 82.67 & 54.60                & 51.15 & 50.38 & 47.35 & 46.96 & 44.94 & 44.57 & 0.697                \\
TRADES \cite{zhang2019theoretically}   &  RN18       & 81.73 & 57.39                & 54.80 & 54.43 & 52.39 & 52.16 & 48.95 & 48.75 & 0.743                \\
TR-GAT (\textbf{Ours})  &  RN18    & 81.32 & 57.61                & 55.34 & 55.13  & 53.37 & 53.22 & 49.77 & 49.62 & \textbf{0.744} \\
TRADES \cite{zhang2019theoretically} $^\dagger$  & WRN34  & 84.92 & 61.06 & 58.47 & 58.09 & 55.79 & 55.56 & 53.29 & 53.18 & 0.705                \\
TR-GAT (\textbf{Ours}) & WRN34& 83.58 & \textbf{61.22}                & \textbf{58.69} & \textbf{58.98} & \textbf{57.07} & \textbf{56.89} & \textbf{53.43} & \textbf{53.32} & 0.719  \\          
\midrule

\end{tabular}}
\vspace{-0.5cm}
\end{table}

\subsection{Evaluation of the proposed defense (GAT)}
The white-box accuracy of the proposed defense GAT is compared with existing defenses in Table-\ref{table:cifar10_whitebox}. In addition to evaluation against standard attacks, we also report accuracy on the recently proposed ensemble of attacks called AutoAttack \cite{croce2020reliable}, which has been successful in bringing down the accuracy of many existing defenses by large margins. The existing single-step defenses are presented in the first partition of the table and the multi-step defenses are presented in the second. The proposed single-step defense GAT outperforms the current state-of-the-art single-step defense, FBF \cite{wong2020fast} on both ResNet-18 \cite{he2016deep} and WideResNet-34-10 \cite{zagoruyko2016wide} models by a significant margin. In fact, we find that increasing model capacity does not result in an increase in robustness for FBF due to catastrophic overfitting. However, with the proposed GAT defense, we obtain a $2.97\%$ increase in worst-case robust accuracy by using a larger capacity model, alongside a significant boost of $4.68\%$ in clean accuracy. In addition to these results, the GAT WRN34-10 model is also evaluated against other state-of-the-art attacks, including our proposed attack GAMA in Table-\ref{table:attacks_main}. Here, GAMA also serves as an adaptive attack to our defense, as the same loss formulation is used for both. We present evaluations on black-box attacks, gradient-free attacks, targeted attacks, untargeted attacks with random restarts and more adaptive attacks Section-\ref{sec:eval_defense} of the Supplementary. We also present all the necessary evaluations to ensure the absence of gradient masking \cite{athalye2018obfuscated} in the Supplementary material. 

We further analyse the impact of using the proposed Guided Adversarial attack for adversary generation in the TRADES training algorithm. We utilize adversaries generated using GAMA-FW for this, as this algorithm generates stronger 10-step attacks when compared to others. Using this approach, we observe marginal improvement over TRADES accuracy. This improves further by replacing the standard adversaries used for TRADES training with GAMA-FW samples only in alternate iterations. We present results on the proposed 10-step defence TR-GAT using this combined approach in Table-\ref{table:cifar10_whitebox}. 

The improvement in robustness with the use of Guided Adversarial attack based adversaries during training is significantly larger in single-step adversarial training when compared to multi-step adversarial training. This is primarily because single-step adversarial training is limited by the strength of the adversaries used during training, while the current bottleneck in multi-step adversarial training methods is the amount of data available for training \cite{carmon2019unlabeled}.

\section{Conclusions}
\label{conclusions}
We propose Guided Adversarial Margin Attack (GAMA), which utilizes the function mapping of clean samples to guide the generation of adversaries, resulting in a stronger attack. We introduce an $\ell_2$ relaxation term for smoothing the loss surface initially, and further reduce the weight of this term gradually over iterations for better optimization. We demonstrate that our attack is consistently stronger than existing attacks across multiple defenses. We further propose to use Frank-Wolfe optimization to achieve faster convergence in attack generation, which results in significantly stronger $10$-step attacks. We utilize the adversaries thus generated to achieve an improvement over the current state-of-the-art adversarial training method TRADES. The proposed Guided Adversarial attack aids the initial steps of optimization significantly, thereby making it suitable for single-step adversarial training. We propose a single-step defense, Guided Adversarial Training (GAT) which uses the proposed $\ell_2$ relaxation term for both attack generation and adversarial training, thereby achieving a significant improvement in robustness over existing single-step adversarial training methods.

\section{Broader Impact}
As Deep Networks see increasing utility in everyday life, it is essential to be cognizant of their worst-case performance and failure modes. Adversarial attacks in particular could have disastrous consequences for safety critical applications such as autonomous navigation, surveillance systems and medical diagnosis. In this paper, we propose a novel adversarial attack method, GAMA, that reliably bounds the worst-case performance of Deep Networks for a relatively small computational budget. We also introduce a complementary adversarial training mechanism, GAT, that produces adversarially robust models while utilising only single-step adversaries that are relatively cheap to generate. Thus, our work has immense potential to have a positive impact on society, by enabling the deployment of adversarially robust Deep Networks that can be trained with minimal computational overhead. During the development phase of systems that use Deep Networks, the GAMA attack can be used to provide reliable worst-case evaluations, helping ensure that systems behave as expected when deployed in real-world settings. On the negative side, a bad-actor could potentially use the proposed attack to compromise Deep Learning systems. However, since the proposed method is a white-box attack, it is applicable only when the entire network architecture and parameters are known to the adversary, which is a relatively rare scenario as model weights are often kept highly confidential in practice.

\section{Acknowledgments and Disclosure of Funding}
This work was supported by Uchhatar Avishkar Yojana (UAY) project (IISC\_10), MHRD, Govt. of India. We would like to extend our gratitude to all the reviewers for their valuable suggestions.

\bibliographystyle{abbrvnat}
\small{\bibliography{references}}

\clearpage

\begin{center}
\textbf{\huge Supplementary Material}
\vspace{0.7cm}
\end{center}
\stepcounter{myequation}
\stepcounter{myalgorithm}
\stepcounter{myfigure}
\stepcounter{mytable}
\stepcounter{mysection}

\makeatletter

\renewcommand{\theequation}{S\arabic{equation}}
\renewcommand{\thefigure}{S\arabic{figure}}
\renewcommand{\thetable}{S\arabic{table}}
\renewcommand{\thesection}{S\arabic{section}}
\renewcommand{\thealgorithm}{S\arabic{algorithm}}

\section{Improved local properties induced by Guided Adversarial Training (GAT)}
\label{sec:local_prop}
In this section, we present details on the improved local properties achieved using the proposed single-step defense, GAT (Guided Adversarial Training). 

We examine the local properties of networks trained using the proposed methodology here. Formally, a function $f$ is locally Lipschitz on a metric space $\mathcal{X}$, if for every $x \in \mathcal{X}$, there exists a neighborhood $U(x)$ such that $f$ restricted to $U(x)$ is Lipschitz continuous, that is, 
\begin{equation}
    \norm{ f(x) - f(y)} \leq \mathcal{L} \cdot \norm{x-y} ~~~~~~~\forall y \in U(x)
\end{equation}

In our framework, we consider $\mathcal{X}$ to be the data-manifold, and $f_{\theta}$ as the softmax output of the neural network, where $\theta$ represents the parameters of the network. 
We first study the impact of the proposed squared $\ell_2$ distance term in the loss function. Minimisation of this regularizer term leads to the following solution for $\theta^*$:
\begin{equation}
    \theta^* = \underset{\theta}{\operatorname{argmin}} {\norm{f_{\theta}(x) - f_{\theta}(\widetilde{x})}}^2_2
\end{equation}
where $\widetilde{x}$ is an adversary corresponding to clean image $x$. We note that the gradient of the loss on $f_\theta(x)$ represents the direction of steepest increase of the loss function. Thus, given that we want to obtain the strongest adversary achievable within a single backward-pass of the loss, we find $\widetilde{x}$ as given in Alg.\ref{alg:GAT}, L\textcolor{red}{6} to L\textcolor{red}{9}.

Since we want the network to be robust to adversaries lying within the $\ell_{\infty}$-ball of radius $\varepsilon$ centered at $x$, we ideally want a function $f_{\theta^*}$ that is locally Lipschitz within $U_{\varepsilon}(x)$:
\begin{equation}
    U_{\varepsilon}(x) = \{x' : \norm{x - x'}_{\infty} \leq \varepsilon \}
\end{equation}
Given that $\norm{x - \widetilde{x}}_{\infty} \leq \varepsilon$, we have,

\begin{equation}
    \norm{x - \widetilde{x}}_{2}  = \sqrt{\sum_{i=1}^{d} (x_i - \widetilde{x}_i)^2} \leq \sqrt{\sum_{i=1}^{d} \varepsilon^2}  \leq \sqrt{d} \cdot \varepsilon
\end{equation}
where $d$ is the dimension of the input space. 
For constrained adversaries, we now have,
\begin{equation}
    \norm{ f_{\theta^*}(x) -  f_{\theta^*}(\widetilde{x})}_2 \leq \sqrt{d}\cdot \varepsilon \cdot \mathcal{L}
\end{equation}
Thus, the prediction of $f_{\theta^*}$ is constant on $U_{\varepsilon}(x)$, if the Lipschitz constant $\mathcal{L}$ is sufficiently small. Note that under this (strong) assumption, $f_{\theta^*}$ is \textit{guaranteed} to be adversarially robust, as it predicts the same class for all images in the $\varepsilon$-constraint.

We note that the set $\mathcal{X}$ is compact, since it is a closed and bounded subset of $[0,1]^d$. Since the function $f_{\theta}$ is differentiable over $\mathcal{X}$, it is uniformly continuous over $\mathcal{X}$ as well. Thus, there exists $\varepsilon' > 0$, such that prediction of $f_{\theta}$ is constant on $U_{\varepsilon'}(x)$ for all $x \in \mathcal{X}$. Thus, since adversarial perturbations cannot lie within an $\varepsilon'$-ball of any sample $x$, we are primarily interested in adversaries $\widetilde{x}$ such that:
\begin{equation}
   \widetilde{x} \in \mathcal{A}(x) =  \{x' : \varepsilon' < \norm{x - x'}_{\infty} \leq \varepsilon , f_{\theta}(x) \neq f_{\theta}(x') \}
\end{equation}
Thus, the local Lipschitz constant $\mathcal{L}$ of interest is given by:
\begin{equation}
   \mathcal{L} = \sup_{\widetilde{x} \in \mathcal{A}(x)} \frac{\norm{ f_{\theta}(x) -  f_{\theta}(\widetilde{x})}_2}{\norm{ x -  \widetilde{x}}_2} < \sup_{\widetilde{x} \in \mathcal{A}(x)} \frac{1}{\sqrt{d} \cdot \varepsilon'} \norm{ f_{\theta}(x) -  f_{\theta}(\widetilde{x})}_2
\end{equation}

The square of the expression on the RHS is precisely the regularisation term used in the proposed loss function for training. Hence, imposing the proposed regularizer encourages the optimization procedure to produce a network that is locally Lipschitz continuous, with a smaller local Lipschitz constant. The actual optimisation procedure minimises the combined loss, with the first term given by the cross-entropy term, and the squared $\ell_2$ loss term weighted by a factor $\lambda$. Note that without the inclusion of the first term, several degenerate solutions are possible, for example, a network that is constant
for all images $x$. The value of $\lambda$ determines the \textit{effective} learning rate for the squared $\ell_2$ loss term, and thus enforces the extent of function smoothness (refer Section-\ref{sec-details on training algo}). The value of $\lambda$ can be chosen so as to achieve the desired trade-off between clean accuracy and robustness \cite{tsipras2018robustness}. For a fixed $\lambda$, we obtain a family of functions $\mathcal{F}_{\lambda}$ that achieve the same cross-entropy loss. We can then extend the same analysis to $f_{\theta^*}$ which is the minimiser of the squared $\ell_2$ loss term, and thus the combined total loss, amongst all functions $f_{\theta} \in \mathcal{F}_{\lambda} $.

\begin{algorithm}[tb]
   \caption{Guided Adversarial Training}
   \label{alg:GAT}
   
\begin{algorithmic}[1]
   \STATE {\bfseries Input:} Network $f_{\theta}$ with parameters $\theta$, Training Data $\mathcal{D} =\{(x_i,y_i) \}$, Minibatch Size M, Attack Size $\varepsilon$, Initial Noise Magnitude $\alpha$, Epochs $E$, Learning Rate $\eta$ 
\FOR{$epoch=1$ {\bfseries to} $E$}
    \FOR{minibatch $B_j \subset \mathcal{D} $}
    \STATE Set $L = 0$
        \FOR{$i=1$ {\bfseries to} $M$}
                
                \STATE $\delta = Bern(-\alpha,\alpha)$   
                \STATE $\delta = \delta +  \varepsilon \cdot$sign$\left(\nabla_{\delta} \left( \ell_{CE}(f_{\theta} (x_i + \delta) , y_i ) + \lambda \cdot || f_{\theta} (x_i+\delta) -  f_{\theta} (x_i)   ||_2^2 \right)\right)$
                
                \STATE $\delta = Clamp~(\delta,-\varepsilon,\varepsilon)$
                \STATE $\widetilde{x}_i = Clamp~(x_i + \delta,0,1)$
                
                \STATE $L = L +  \ell_{CE}(f_{\theta} (x_i) , y_i ) + \lambda \cdot || f_{\theta} (\widetilde{x}_i) -  f_{\theta} (x_i) ||_2^2$

        \ENDFOR
        
        \STATE $\theta = \theta - \frac{1}{M} \cdot \eta \cdot \nabla_{\theta} L   $
    \ENDFOR
\ENDFOR
\end{algorithmic}
\end{algorithm}
\vspace{-0.3cm}

\section{Details on the datasets used}
\label{sec:datasets}

We run extensive evaluations on MNIST \cite{lecun1998mnist}, CIFAR-$10$ \cite{krizhevsky2009learning} and ImageNet \cite{imagenet_cvpr09} datasets to validate our claims on the proposed attack and defense.  

MNIST \cite{lecun1998mnist} is a handwritten digit recognition dataset consisting of 60,000 training images and 10,000 test images. The images are grayscale, and of dimension 28$\times$28. We split the training set into a random subset of 50,000 training images and 10,000 validation images.

CIFAR-$10$ \cite{krizhevsky2009learning} is a popular dataset in computer vision research, consisting of the following ten classes: Airplane, Automobile, Bird, Cat, Deer, Dog, Frog, Horse, Ship and Truck. The similarity of classes such as Cat and Dog make this a challenging dataset for the domain of adversarial robustness. The dimension of each image in this dataset is $32\times32\times3$. The original training set comprises of 50,000 images which we split into 49,000 training images and 1,000 validation images (equally balanced across all ten classes), while the test set has 10,000 images.

ImageNet \cite{imagenet_cvpr09} is a 1000-class dataset consisting of approximately 1.2 million training images and 50,000 images in the validation set. This dataset has a private test set which is not available for access to the public. Therefore, we use the designated validation set as the test set for our experiments. Furthermore, we split the designated training set into an 80-20 train-validation split. For training and evaluation of our proposed defense, we consider a random 100-class subset of this dataset, in order to ease the computation time and resource requirements. Even this subset is challenging due to the large dimensionality of the input space ($224\times224\times3$) and the high level of similarity between different classes. The set of $100$ classes used for our experiments is shared along with our codes. 

We use NVIDIA DGX workstation with V100 GPUs for our training and evaluations. The proposed single-step defense takes approximately $2$ hours for training on ResNet-$18$ architecture for CIFAR-$10$ dataset. The proposed 100-step GAMA-PGD attack takes approximately $5$ minutes for a single run to evaluate a ResNet-18 model on the CIFAR-$10$ test set. 

\section{Details on Guided Adversarial Training}
\label{sec:GAT}

In this section, we present implementation details of the proposed defense. 

\subsection{Architecture details} 
\begin{table}
\caption{Architecture used for MNIST dataset. Modified LeNet architecture is used for training robust models. We use the network, BB-MNIST as the source model for Black-Box attacks on the MNIST dataset}
\centering
\label{table:mnist_architecure}
\setlength\tabcolsep{8pt}{
    \begin{tabular}{@{}|c|c|@{}} 
    \toprule
    \textbf{Modified LeNet (M-LeNet)} & \textbf{BB-MNIST} \\ \midrule
    \{conv(32,5,5) + Relu\}$\times$2 & Conv(64,5,5) + Relu \\
    MaxPool(2,2) & Conv(64,5,5) + Relu  \\
    \{conv(64,5,5) + Relu\}$\times$2 & Dropout(0.25)         \\
    MaxPool(2,2) & FC(128) + Relu  \\ 
    FC(512) + Relu & Dropout(0.5) \\
    FC + Softmax    & FC + Softmax \\
    \bottomrule
    \end{tabular}}
\end{table}

For evaluation of the proposed defense, we select a fixed architecture for each dataset, and use the same architecture to report results across all existing defense methods as well. We use a modified LeNet architecture with 4 convolutional layers as shown in Table-\ref{table:mnist_architecure} for MNIST, and the ResNet-18 \cite{he2016deep} architecture for our experiments on CIFAR-$10$ and ImageNet-$100$ datasets. We also report results on  WideResNet-34-10 \cite{zagoruyko2016wide} architecture for the CIFAR-$10$ dataset with the proposed defense.

\subsection{Details on the training algorithm}
\label{sec-details on training algo}
The training algorithm for the proposed single-step defense GAT is presented in Algorithm-\ref{alg:GAT}. This is explained in Section-\ref{main_sec:GAT} of the main paper.
In the proposed defense GAT, we first add Bernoulli noise of magnitude $\alpha$ to the clean image. We set the value of $\alpha$ to be either $\varepsilon$ or $\varepsilon$/2 \cite{tramer2017ensemble}. In the next step, we generate a GA-CE (Guided Adversarial Cross-Entropy) attack on the noise-added image, and finally project the generated perturbation onto the $\varepsilon$-ball of the clean image. In standard single-step adversarial training \cite{tramer2017ensemble}, the  FGSM attack is of magnitude ($\varepsilon$ - $\alpha$). So, the perturbation is always within the $\ell_{\infty}$ $\varepsilon$-ball of the clean image and thus there is no need to project it back to the $\varepsilon$-ball. The proposed formulation however, can move the adversary outside the $\varepsilon$-ball thereby increasing the likelihood of the adversary lying on the boundary of the $\varepsilon$-ball after projection. In principle, this broader class of adversaries should lead to a stronger attack, since most of the adversaries are farther away from the original image when compared to a standard R-FGSM attack. 

The two losses in the proposed framework (Alg.\ref{alg:GAT}, L\textcolor{red}{7},\textcolor{red}{10}) are combined by weighting the squared $\ell_2$ loss term by a factor $\lambda$. This weighting term determines the trade-off between accuracy and robustness \cite{tsipras2018robustness} as shown in Fig.\ref{fig:acc_vs_hyp}(b). We observe that during the initial stages of training, the loss surface is relatively smooth, thereby requiring a low value for $\lambda$. This factor is stepped up towards the end of training. 
The learning-rate is decayed when the loss begins to plateau. We generally observe that for the first learning-rate update, both clean and adversarial accuracy improve in tandem if $\lambda$ is kept fixed. As training progresses, the loss surface becomes increasingly convoluted; we thus step-up $\lambda$ along with the subsequent decay in learning rate, so as to strike a balance between clean accuracy and adversarial robustness. 

\begin{figure}
        \includegraphics[width=\linewidth]{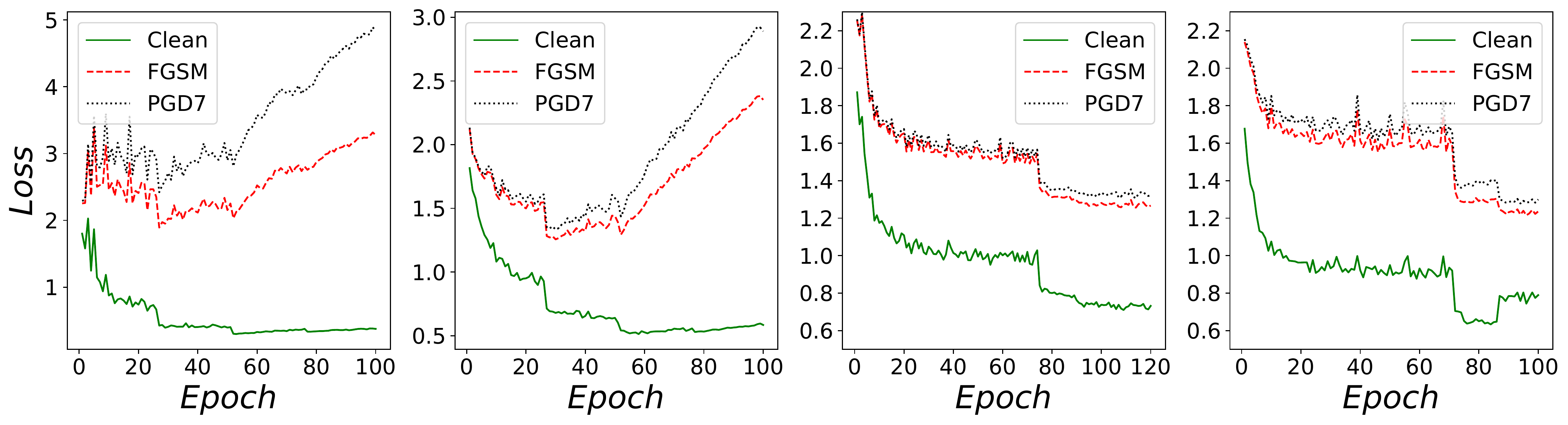}
        \vspace{-0.3cm}
        \caption{Plot of Cross-Entropy loss on CIFAR-10 test samples across epochs for different training methods: (a) R-FGSM-AT (b) PGD-AT (c) TRADES (d) GAT (\textbf{Ours}) }
        
        \label{fig:lossvsepoch}
        \vspace{-0.4cm}
\end{figure}

The $\lambda$ step-up factor can be viewed as an effective learning rate increase for the squared $\ell_2$ loss term. During adversarial training involving step learning rate decay, accuracy is boosted significantly for the initial few step decays. This results in a change in loss landscape, leading to an increase in loss on adversarial samples, as can be seen in case of R-FGSM training and PGD training in Fig. \ref{fig:lossvsepoch} (a and b). The training on adversarial samples is however unable to compensate for the increase in loss as the learning rate is too low. Thus, inclusion of the step-up factor ensures that the adversarial loss does not increase rapidly over epochs. It can be seen in  Fig.\ref{fig:lossvsepoch} (d) that a combination of the proposed learning rate schedule and step-up factor is able to prevent an increase in loss. Similar to TRADES, the loss on clean and adversarial samples consistently reduces over epochs in the proposed method. It is also worth noting that the loss on FGSM samples is very close to the loss on PGD samples, thereby proving the effectiveness of training with single-step adversaries generated using the Guided Adversarial attack.

\subsection{Implementation details}
\label{sec-implementation details}
\begin{figure}
\centering
        \includegraphics[width=\linewidth]{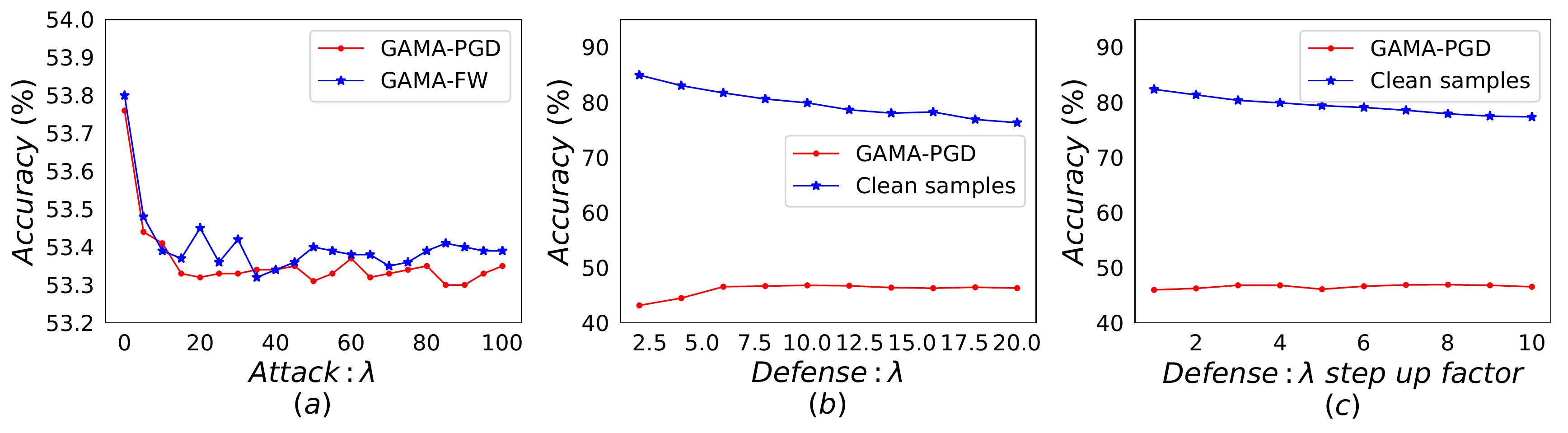}
        \vspace{-0.6cm}
        \caption{\small{(a) Accuracy ($\%$) of the TRADES-WRN34 model \cite{zhang2019theoretically} trained on CIFAR-10 dataset, against the proposed GAMA-PGD and GAMA-FW attacks, across various settings of $\lambda$ in Eq.\ref{Attack_loss} of the main paper. This is the weight of $\ell_2$ relaxation term in the loss. (b, c) Accuracy ($\%$) of the proposed GAT defense on ResNet-$18$ models trained on CIFAR-10 dataset, across variation in hyperparameters used for the defense. $\lambda$ (Alg.-\ref{alg:GAT}, L\textcolor{red}{7}, \textcolor{red}{10}) is varied in (b) and $\lambda$ step up factor is varied in (c). Accuracy ($\%$) on clean samples and 100-step GAMA-PGD  adversaries is shown. GAMA-PGD attack settings are fixed to optimal values.}
        \label{fig:acc_vs_hyp}}
\end{figure}

\textbf{GAT (Single-step defense):} Before generating the attack for adversarial training in GAT defense, we add initial noise of magnitude $\varepsilon$ for MNIST, and $\varepsilon/2$ for the other datasets. For the CIFAR-$10$ dataset, we use an initial $\lambda$ and step-up factor of $10$ and $4$ respectively for ResNet-$18$ training, and $3$ and $20$ respectively for WideResNet-$34$-$10$ training. The same values of $\lambda$ and step-up factor are used for both generation of attack (Alg.\ref{alg:GAT}, L\textcolor{red}{7}) and training (Alg.\ref{alg:GAT}, L\textcolor{red}{10}). We use the SGD optimizer with  momentum of 0.9 and weight decay of 5e-4 for all our experiments. The learning rate is set to 0.1 and decayed by a factor of $10$, at epochs $70$ and $85$ for ResNet-$18$ and at epochs $55$,$70$ and $75$ for WideResNet-$34$-$10$. As discussed in Section-\ref{sec-details on training algo}, the second learning rate update is accompanied by a $\lambda$ step-up. The impact of variation in $\lambda$ and $\lambda$ step-up factor is shown in Fig.\ref{fig:acc_vs_hyp}(b) and Fig.\ref{fig:acc_vs_hyp}(c) respectively. It can be observed that the clean accuracy and accuracy on GAMA-PGD adversarial samples is stable across variations in the hyperparameters. As $\lambda$ increases, there is a reduction in clean accuracy and an increase in robustness. This trend continues till a value of $10$, after which the adversarial accuracy remains constant or starts reducing. We therefore select $10$ as the optimum value. A similar trend is observed for variation in $\lambda$ step-up factor as well. The slight reduction in robustness at high values of $\lambda$ or $\lambda$ step-up factor is due to the reduction in clean accuracy. Since accuracy on clean samples is an upper bound on adversarial robustness, over-regularization causes both to reduce. 

For ImageNet-$100$, we use an initial $\lambda$ of $20$ and step it up by a factor of $7$ at epoch $90$. The learning rate is 0.1 initially and decayed by a factor of $10$ at epochs $60$ and $90$.

For MNIST, the initial learning rate is set to $0.01$, and further decayed by a factor of $5$ three times at regular intervals. In MNIST dataset, the clean accuracy shoots up to above $90$\% within the first epoch, and reaches a very high value in a few epochs. Hence, we do not need a simple drop in learning rate without $\lambda$ step-up, for a further increase in robust accuracy. We therefore include the $\lambda$ step-up factor at all three times of learning rate update. We set the value of $\lambda$ to 
$15$ and the $\lambda$ step-up factor to $3$.

We train our MNIST model for 50 epochs, CIFAR-$10$ model for 100 epochs and ImageNet-$100$ model for 120 epochs. The other methods are trained until convergence.

\textbf{TR-GAT (Multi-step defense):} We next present details on the proposed multi-step defense TR-GAT. To incorporate Guided Adversarial Attacks for TRADES training as shown in Table-\ref{table:cifar10_whitebox} in the main paper, we alternate between a 10-step PGD attack that maximises KL-Divergence and a 10-step GAMA-FW attack with a constant $\lambda$, set to the same value as the weighting used for the KL-Divergence term in training. We set this weighting term (called $\beta$ in the original TRADES paper \cite{zhang2019theoretically}) to be 5 and 6 respectively for the ResNet-18 and WideResNet-34 models on CIFAR-$10$ dataset. For training the TR-GAT model, we use the same learning rate schedule and total epochs as used for the corresponding TRADES model.

\section{Implementation details of the Guided Adversarial Margin Attack}
\label{sec:impl_gama}
The loss function that is maximized for generation of our proposed attack GAMA is shown in Eq.\ref{Attack_loss} of the main paper. This consists of two terms, maximum margin loss and squared $\ell_2$ relaxation term between the softmax vectors of clean and perturbed images. The squared $\ell_2$ term is weighted by a factor $\lambda$, which is decayed to $0$ over a fixed number of iterations. We set $\lambda$ to $50$ and decay this to $0$ over $25$ iterations. This aids the optimization process by providing a better initial direction. As shown in Fig.\ref{fig:acc_vs_hyp}(a), the attack strength is stable over a wide range of $\lambda$ values. 

Prior to the attack generation, we add Bernoulli noise of magnitude $\varepsilon$ to the image. Analogous to the training of Deep Neural Networks, generation of standard attacks are also known to benefit with a step learning rate schedule over the optimization process \cite{gowal2019alternative}. For the GAMA-PGD attack, we use an initial step size of $2\cdot\varepsilon$ and decay it by a factor of $10$ at iterations $60$ and $85$ for a $100$-step schedule. Similarly, for the GAMA-FW attack, we use an initial $\gamma$ of 0.5 and decay it by a factor of $5$ at the same iterations. For evaluating the TRADES leaderboard \cite{zhang2019theoretically} WideResNet-34 CIFAR-10 model, we use a multi-targeted version of our attack; we run the attack for 100 steps, and 20 random restarts, wherein we alternate between the proposed GAMA loss and the margin loss corresponding to different classes over multiple restarts.

As noted by Gowal et al. \cite{gowal2019alternative}, the loss surface of models adversarially trained on the MNIST datasets is complex. This necessitates different attack settings for this dataset. The threat model considered typically for MNIST is $\varepsilon=0.3$. We set the initial $\lambda$ to $5$ and decay it to $0$ in $50$ iterations. For GAMA-PGD, we use an initial step size of $\varepsilon$ and decay it by $10$ at iterations $50$ and $75$. For GAMA-FW, $\gamma$ of $0.5$ is used initially and decayed by a factor of $5$ at the same iterations.

\section{Details on Evaluation of the proposed attack}
\label{sec:eval_attack}
\begin{table}[t]
\caption{\textbf{Ablations (CIFAR-10)}: Accuracy ($\%$) of various attacks (rows) on TRADES-WRN34 \cite{zhang2019theoretically} model under the $\ell_\infty$ bound with $\varepsilon=0.031$, across different settings of number of steps and restarts.}
\setlength\tabcolsep{3pt}
\resizebox{1.0\linewidth}{!}{
\label{table:ablations}
\begin{tabular}{l|cc|cc}
\toprule
                                 \textbf{Attacks}            & \multicolumn{2}{c|}{\textbf{100 - step attacks}} & \multicolumn{2}{c}{\textbf{10 - step attacks}} \\
                                             & \textbf{~~~Single run~~~}          & \textbf{~~5 restarts~~}          & \textbf{~~~Single run~~~}          & \textbf{~~5 restarts~~}         \\
                                             \midrule
PGD (Cross-entropy loss)                     & 55.61                  & 55.27                  & 56.7                   & 56.31                 \\
PGD (Margin loss in logits space)            & 54.19                  & 54.07                  & 55.04                  & 54.75                 \\
PGD (Margin loss in prob. space)             & 53.94                  & 53.8                   & 54.87                  & 54.68                 \\
PGD (Margin loss and $\ell_2$ loss in prob. space) & 53.73                  & 53.54                  & 54.96                  & 54.67                 \\
GAMA - PGD                                   & \textbf{53.29}                  & \textbf{53.17}                  &     54.95                 &          54.66           \\
GAMA - FW                                    & 53.38                  & 53.22                  & \textbf{54.27}                  & \textbf{54.00}   \\
\midrule 
\end{tabular}}
\end{table}

For evaluation of the proposed attacks (Table-\ref{table:attacks_main} and Fig.\ref{fig:acc_vs_rr} of the main paper, Fig.\ref{fig:acc_vs_hyp}(a) of the supplementary), we use pre-trained models shared by the respective authors of various defenses. Therefore, the architecture of different models would be as chosen by the respective authors, and is thus not consistent across all defenses presented in Table-\ref{table:attacks_main} of the main paper. 

\subsection{Ablation Experiments} We present evaluations on the TRADES WideResNet-34 model on the CIFAR-$10$ test set with several ablations of the proposed attack in Table-\ref{table:ablations}. We first observe that the maximum-margin loss, which is similar to the C\&W $\ell_\infty$ based attack \cite{carlini2017towards}, is more effective when compared to the cross-entropy loss, for both $10$ and $100$ step attacks. Further, we observe that we obtain stronger adversaries while optimising the margin loss between predicted probability scores, as compared to the corresponding logits. The weighting factor for the squared $\ell_2$ relaxation term is linearly decreased to 0 for the $100$-step attack, while it is kept constant for the $10$-step attack. From the $100$-step evaluations, we observe that graduated optimisation indeed aids in finding stronger adversaries. Further, the addition of initial Bernoulli random noise aids in improving $100$-step adversaries. We also note that GAMA-FW achieves the strongest attack when the available budget on the number of steps for attack is relatively small, making it suitable for use in multi-step adversarial training. 

\subsection{Variation of Accuracy and Cross-Entropy loss across attack iterations} 
\begin{figure}[t]
        \includegraphics[width=\linewidth]{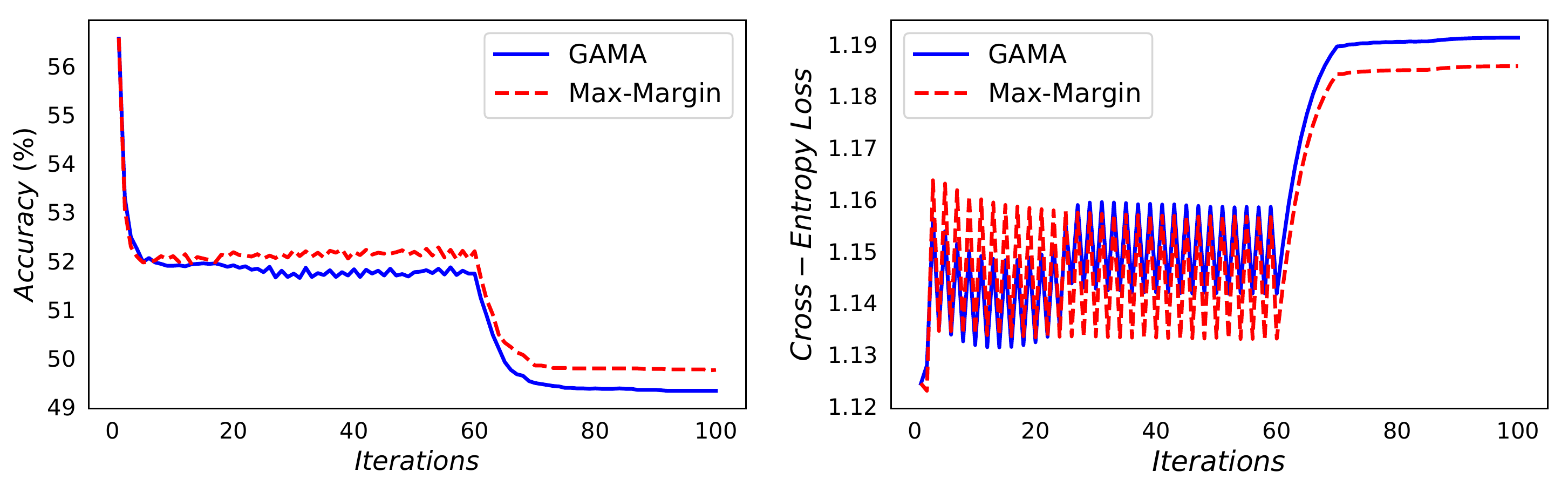}
        \vspace{-0.3cm}
        \caption{\textbf{Variation of Accuracy and Cross-Entropy Loss across attack iterations:} Plot of the Accuracy and Cross-Entropy Loss across attack iterations for the TRADES ResNet-$18$ defense, trained on the CIFAR-$10$ dataset. The proposed attack GAMA (Maximum-margin + $\ell_2$ relaxation term) has been compared with a standard maximum-margin based attack. }
        \label{ce_iterations}
        \vspace{-0.3cm}
\end{figure}

In Fig.\ref{ce_iterations}, we plot the accuracy and Cross-Entropy loss across attack iterations for the TRADES ResNet-18 model on the CIFAR-10 dataset. The GAMA attack achieves lower accuracy and higher Cross-Entropy loss during the course of optimization, as compared to the attack generated using only the maximum-margin loss. We note from Fig.\ref{ce_iterations}(b) that the decay of $\ell_2$ relaxation term over the first 25 iterations in GAMA is crucial to allow Cross-Entropy loss to increase. Therefore, while the $\ell_2$ relaxation term gives the right initialization, switching to the true maximum-margin optimization objective is important for achieving a stronger attack. We thus observe that the additional $\ell_2$ relaxation term with a decaying coefficient indeed aids in the optimization process, and prevents the attack from stalling at points where the primary objective function attains a local maximum. Lastly, although the loss tends to oscillate before the first drop in step-size at iteration $60$, we find that it is important to allow the attack to adequately explore the constraint set, in order to identify strong adversarial perturbations towards the end of optimization.

\subsection{Use of ADAM optimizer in the GAMA attack} We also implement an ablation of the proposed GAMA-PGD attack using the ADAM optimizer \cite{kingma2014adam} with the true gradients, instead of using Stochastic Gradient Descent with signed gradients. We perform a hyperparameter search over the initial step-size, step-schedule and decaying coefficient of the $\ell_2$ smoothing term. We find that the attack is marginally weaker when the ADAM optimizer is used; the strongest 100-step attack obtained over the entire hyperparameter search achieves 53.66\% accuracy on the TRADES WideResNet-34 model for a single run of the attack, compared to 53.29\% as obtained by the original GAMA-PGD attack.

\section{Details on Evaluation of the proposed defense}
\label{sec:eval_defense}
In this section, we present additional experimental results to support our claims on the proposed single-step defense GAT. 
For CIFAR-10 dataset, we report results on ResNet-18 and WideResNet-34-10 architectures for the proposed method in the main paper. We note that while defense methods such as CURE \cite{moosavi2019robustness} and 2-step LLR \cite{qin2019adversarial} achieve non-trivial robustness against multi-step adversaries using adversarial training on two or three step attacks, they are significantly weaker than recent single-step defense methods such as FBF \cite{wong2020fast} and R-MGM \cite{baburaj2019regularizer}. Thus, we restrict our primary comparisons to the latter defenses which are more robust for a similar computational budget. We use the GAT defense trained on the ResNet-18 architecture for further evaluations in this section.

\subsection{Ablation Experiments}
\begin{table}[t]
\caption{\textbf{GAT Ablations (CIFAR-10)}: Accuracy ($\%$) of various ablations of the GAT defense (rows) trained on ResNet-18 model under the $\ell_\infty$ threat model of $\varepsilon=8/255$.}
\setlength\tabcolsep{3pt}
\resizebox{1.0\linewidth}{!}{
\label{table:ablations_defense}
\begin{tabular}{l|ccc}
\toprule
                                           \multicolumn{1}{l}{\textbf{Ablations}}                  & \multicolumn{1}{|c}{\textbf{Clean}} & \multicolumn{1}{c}{\textbf{PGD-100}} & \multicolumn{1}{c}{\textbf{AA}} \\
\midrule
GAT (Proposed method)                                       & 80.49                                  & 53.08                                & 47.30                           \\
\textcolor{black}{\textbf{A1:}} GAT, without alternating between CE and GA-CE attacks     & 80.22                                  & 51.50                                & 46.52                           \\
\textcolor{black}{\textbf{A2:}} GAMA max-margin loss for attack and training                           & 23.22                                  & 15.64                                & 10.98                           \\
\textcolor{black}{\textbf{A3:}} GA-CE attack + standard defense (training on CE$_{clean}$ $+$ CE$_{adv}$) & 90.21                                  & 33.57                                & 32.29                            \\
\textcolor{black}{\textbf{A4:}} Standard attack + standard defense (R-FGSM training)        & 89.24                                  & 34.23                                & 33.16                           \\
\textcolor{black}{\textbf{A5:}} Standard (CE) attack + GAT defense                          & 80.05                                  & 51.8                                & 44.21    \\
\midrule 
\end{tabular}}
\end{table}

We present ablations on the proposed defense GAT, trained on CIFAR-10 dataset in Table-\ref{table:ablations_defense}. The architecture of the models is ResNet-18, and the models are trained to be robust under an $\ell_{\infty}$ threat model of $8/255$. We present results against PGD-100 step attack and the recently proposed ensemble of attacks, AutoAttack \cite{croce2020reliable}. In order to diversify the attacks generated for GAT training, we switch between standard cross-entropy loss and the proposed GA-CE loss (Algo.-\ref{alg:GAT}, L\textcolor{red}{7}) in alternate iterations. However, even without this additional diversification step (Ablation-\textcolor{black}{\textbf{A1}}), we observe similar accuracy on AA with merely a marginal drop. We observe that using the original GAMA loss (Eq.\ref{Attack_loss}) for both generation of attack and adversarial training does not lead to improved robustness (Ablation-\textcolor{black}{\textbf{A2}}). This is because minimization of maximum-margin objective is not suitable for training Deep Neural Networks. 

The proposed defense involves the use of a modified loss function for both attack generation and adversarial training. We perform experiments to evaluate the impact of each of these components individually. The model in Ablation-\textcolor{black}{\textbf{A3}} is trained using GA-CE attack based adversaries. Adversarial training in this experiment in done by minimizing cross-entropy loss on both clean and adversarial samples, in similar vein to R-FGSM adversarial training. While R-FGSM training (Ablation-\textcolor{black}{\textbf{A4}}) leads to an improvement over this method, it is still significantly weaker than the proposed defense. Similarly, we find that using the GAT loss for defense alone (Ablation-\textcolor{black}{\textbf{A5}}) does not lead to significantly improved robustness. Therefore a combined usage of the loss in both attack generation and adversarial training is crucial for the state-of-the-art results obtained using GAT.

\subsection{Stability of Guided Adversarial Training}
In this section, we investigate the stability of the proposed training algorithm on the CIFAR-$10$ dataset. We train a ResNet-18 model multiple times allowing different random initialisation of network parameters in each run. For each run, we follow the training methodology as outlined in Sections \ref{sec-details on training algo} and \ref{sec-implementation details}. We observe that models trained using GAT are very stable; the PGD-100 accuracy obtained over six random  reruns are as follows:  52.14, 51.7, 52.02, 52.35, 51.96, 51.74. The low variance (Standard Deviation = 0.224) across multiple runs highlights the stability of the proposed training method. Further, we note that models trained using GAT do not suffer from catastrophic overfitting, as observed in prior works such as FBF \cite{wong2020fast}. In Fig.\ref{fig:lossvsepoch}, we observe that the Cross-Entropy loss on adversaries generated using an FGSM attack is highly similar to the loss on PGD 7-step adversaries throughout the entire training regime, indicating the absence of catastrophic overfitting. We also observe that even the model obtained in the last epoch of training achieves high adversarial accuracy against strong multi-step attacks, in sharp contrast to models obtained towards the end of training using FBF.

\subsection{Loss Surface Plots}

\begin{figure*}[t]
        \includegraphics[width=\linewidth]{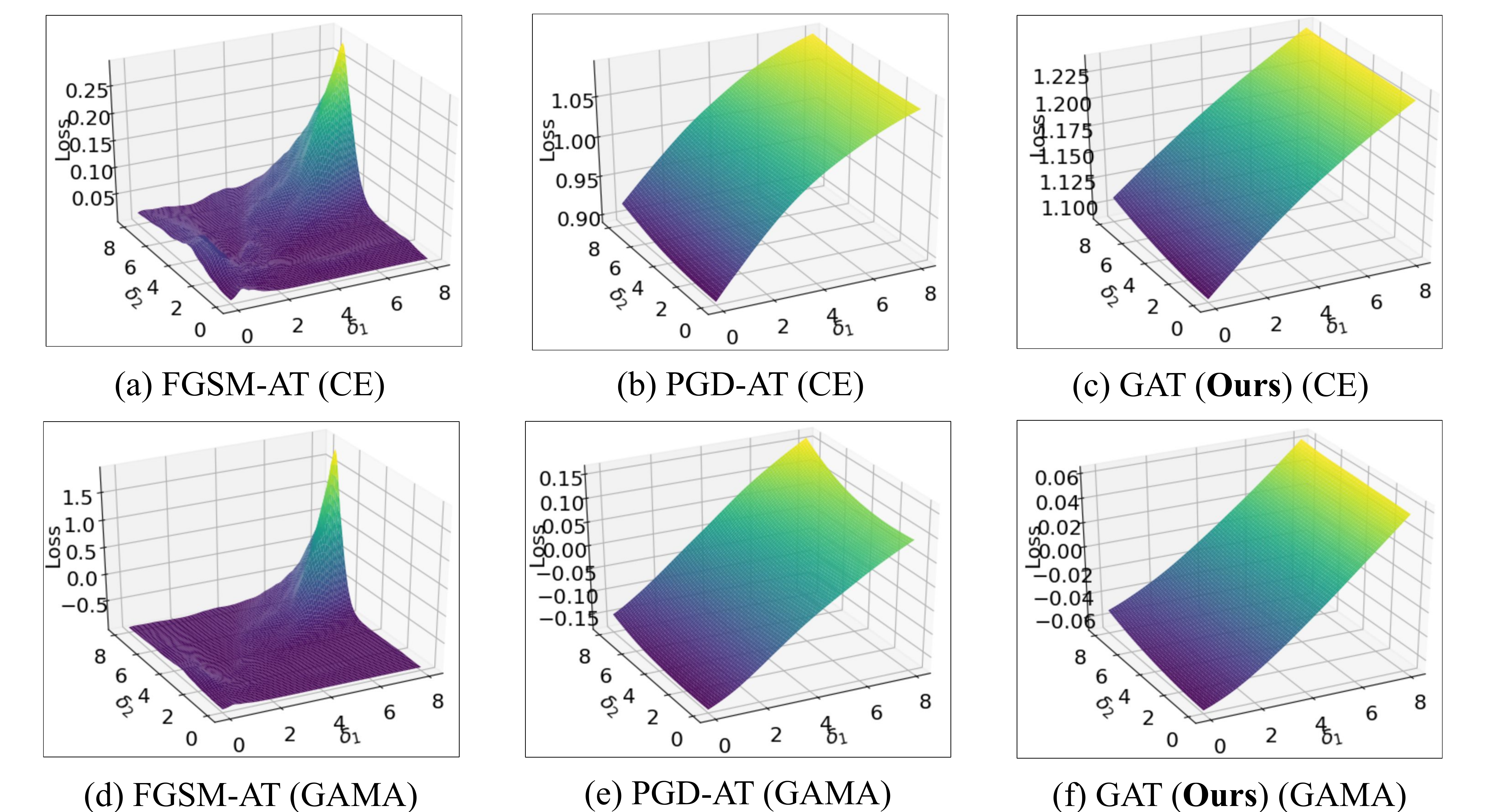}
        \vspace{-0.3cm}
        \caption{Plot of the loss surface of different models on perturbed images of the form $x^* = x+\delta_1 g+\delta_2 g^\perp$, obtained by varying $\delta_1$ and $\delta_2$. Here $g$ is the sign of the gradient direction of the cross-entropy loss with respect to the clean image ($x$) and $g^\perp$ is a direction orthogonal to $g$.}
        \label{loss_surface}
        \vspace{-0.3cm}
\end{figure*}

To verify the absence of gradient masking, we visualise the loss surface of models trained using the proposed single-step defense GAT, in the neighbourhood of a test data sample. To generate the loss surface, we plot the loss obtained by perturbing the clean sample $x$ along two directions: one along the direction of the gradient of the loss at sample $x$, and another direction that is orthogonal to the gradient. In Fig.\ref{loss_surface} (a), (b) and (c), we plot the standard cross-entropy loss for FGSM-AT, PGD-AT and GAT trained models respectively. We find that FGSM training produces models with significant gradient masking and a convoluted loss surface. On the other hand, for PGD-AT and GAT models, the loss surface is smooth, thereby verifying the absence of gradient masking in the proposed single-step defense. In the second row of Fig.\ref{loss_surface}, we plot the proposed GAMA loss (Eq.\ref{Attack_loss} of the main paper), which is a combination of the maximum margin loss and the squared $\ell_2$ distance between the softmax predictions of the perturbed image and original data sample respectively. The squared $\ell_2$ relaxation term is weighted by a constant value of $\lambda = 25$ for obtaining the loss surface plots in Fig.\ref{loss_surface} (d), (e) and (f). We again find that the loss surface for the proposed defense GAT is smooth, despite being a single-step defense method. We further note that the GAMA loss surface is smoother than the cross-entropy loss surface for all three models, specifically for the FGSM model. This helps explain why the proposed GAMA attack is more effective than the standard maximisation of the cross-entropy loss.

\subsection{Performance against White-Box attacks}
\begin{table}[t]
\caption{\textbf{Defenses (ImageNet-100)}: Accuracy ($\%$) of different defenses (rows) trained on ResNet-18 architecture against various $\ell_\infty$ norm bound ($8/255$) white-box attacks (columns). The first partition corresponds to single-step defenses, and the second to multi-step defenses. For the Carlini and Wagner (C\&W) attack, the Mean-$\ell_2$ norm required to achieve high Fooling Rate (FR) is reported. Higher the $\ell_2$ norm, better is the robustness.}
\setlength\tabcolsep{7pt}
\resizebox{1.0\linewidth}{!}{
\label{table:imagenet_whitebox}
\begin{tabular}{l|c|c|c|ccc|c|c|c}
\toprule
&\textbf{Clean~} & \multicolumn{7}{c|}{\textbf{Acc (\%) on $\ell_{\infty}$ attacks}} &\multicolumn{1}{|c}{\textbf{C \& W}}\\ 
\textbf{Method} &  \textbf{Acc (\%)} & \textbf{FGSM}  &\textbf{IFGSM}& \multicolumn{3}{c|}{\textbf{PGD (n-steps)}}&\textbf{GAMA}&\textbf{AA} & \textbf{Mean}        \\
                     &         &   & \textbf{7-step}    &\textbf{7}    & \textbf{20} & \textbf{500} & \textbf{PGD-100}& &\textbf{$\ell_2$} \\ \midrule
Normal   & \textbf{81.44} & 8.22  & 0.08  & 0.06  & 0.02  & 0.00  & 0.00 & 0.00 & 0.570  \\
RFGSM-AT \cite{tramer2017ensemble}& 78.46 & 32.04 & 23.52 & 21.64 & 15.86 & 13.88 & 13.38 &12.96 & \textbf{2.960}  \\
FBF \cite{wong2020fast}     & 57.32 & 36.24 & 26.92 & 29.84 & 28.00 & 27.22 & 21.78&20.66 & 0.737 \\
R-MGM \cite{baburaj2019regularizer}   & 64.84 & 40.80 & 35.18 & 35.60 & 32.48 & 31.68 & 27.46&27.68 & 1.636 \\

GAT (\textbf{Ours})  & 67.98 & \textbf{45.38} & \textbf{39.66} & \textbf{40.18} & \textbf{38.02} & \textbf{37.46} &\textbf{29.30}& \textbf{28.92} & 1.499 \\
\midrule
PGD-AT \cite{madry-iclr-2018}  & 68.62 & 43.04 & 40.00 & 39.64 & 37.20 & 36.56 & 32.24 &32.98 & 1.550  \\
TRADES \cite{zhang2019theoretically} & 62.88 & 40.46 & 38.52 & 38.44 & 37.34 & 37.24&31.44&31.66 &   1.360  \\
\midrule
\end{tabular}}
\end{table}

\begin{table}[t]
\caption{\textbf{Defenses (MNIST)}: Accuracy ($\%$) of different defenses (rows) trained on M-LeNet architecture (Table-\ref{table:mnist_architecure}) against various $\ell_\infty$ norm bound ($0.3$) white-box attacks (columns). The first partition corresponds to single-step defenses, and the second to multi-step defenses. For the Carlini and Wagner (C\&W) attack, the Mean-$\ell_2$ norm required to achieve high Fooling Rate (FR) is reported. Higher the $\ell_2$ norm, better is the robustness.}
\setlength\tabcolsep{7pt}
\resizebox{1.0\linewidth}{!}{
\label{table:mnist_whitebox}
\begin{tabular}{l|c|c|c|ccc|c|c|c}
\toprule
&\textbf{Clean~} & \multicolumn{7}{c|}{\textbf{Acc (\%) on $\ell_{\infty}$ attacks}} &\multicolumn{1}{|c}{\textbf{C \& W}}\\ 
\textbf{Method} &  \textbf{Acc (\%)} & \textbf{FGSM}  &\textbf{IFGSM}& \multicolumn{3}{c|}{\textbf{PGD (n-steps)}}&\textbf{GAMA}&\textbf{AA} & \textbf{Mean}        \\
                     &         &   & \textbf{40-step}    &\textbf{40}    & \textbf{100} & \textbf{500} & \textbf{PGD-100}& &\textbf{$\ell_2$} \\ \midrule
Normal   & 99.20                      & 16.59                      & 0.48                       & 0.02                       & 0.00                       & 0.00                       & 0.00                       & 0.00                       & 1.42 \\
RFGSM-AT \cite{tramer2017ensemble}& \textbf{99.37}                      & 92.44 & 89.47 & 90.24 & 85.85 & 85.32                      & 83.64                      & 82.28                      & 2.19 \\
FBF \cite{wong2020fast}     & 99.30 & \textbf{97.47}                      & 94.53                      & 94.85                      & 92.35                      & 91.37                      & 87.27                      & 79.02                      & 1.91 \\
R-MGM \cite{baburaj2019regularizer}   & 99.04 & 96.35 & 93.09                      & 93.06 & 90.96 & 90.56                      & 88.13                      & 86.21                      & \textbf{2.31} \\

GAT (\textbf{Ours})   & \textbf{99.37} & 97.11                      & \textbf{95.61}                      & \textbf{96.11} & \textbf{94.58} & \textbf{94.44} & \textbf{92.96} & \textbf{90.62} & 2.30 \\
\midrule
PGD-AT \cite{madry-iclr-2018}  & 99.27                      & 96.27 & 94.91 & 95.53 & 94.14 & 93.98                      & 92.80                      & 91.81                      & 2.63 \\
TRADES \cite{zhang2019theoretically}  & 99.32                      & 96.08                      & 94.86                      & 95.26                      & 93.52                      & 93.40                      & 92.74                      & 92.19                      & 2.53 \\
\midrule
\end{tabular}}
\end{table}

The results on white-box adversarial attacks for ImageNet-$100$ and MNIST datasets are presented in Table-\ref{table:imagenet_whitebox} and Table-\ref{table:mnist_whitebox} respectively. On both datasets, we observe significant improvement in robustness with the proposed approach when compared to existing single-step adversarial training methods. We also note that the robustness achieved is comparable to the multi-step adversarial training methods, TRADES and PGD-AT, presented in the second partition of both tables.

We also evaluate all the defenses on MNIST and ImageNet-100 datasets against the proposed GAMA-PGD attack. The proposed defense is stronger compared to all other defenses even on the GAMA-PGD 100-step attack. The proposed GAMA-PGD attack is notably stronger than all the single attacks considered here. We note that for ImageNet-100 dataset, a single run of the GAMA-PGD attack is comparable to the AA attack, which is an ensemble of multiple attacks with five random restarts each. In particular, the GAMA-PGD attack is significantly stronger than the APGD-CE, APGD-DLR, FAB and Square attacks that constitute the AA ensemble attack. 
For MNIST dataset, GAMA-PGD is comparable to AA on some defenses and significantly weaker than AA on few others. This is primarily because one of the attacks in the AA ensemble is the Square attack, which is a query based attack. This is a gradient-free attack and is therefore significantly stronger than gradient-based attacks in cases where the loss surface is complex, leading to masking of the true gradient direction.

While the defense is trained to be robust against $\ell_{\infty}$ norm bound perturbations, we find that the robustness to the $\ell_2$ norm based Carlini \& Wagner (C\&W) attack \cite{carlini2017towards} is comparable to other $\ell_\infty$ norm based adversarial training methods. We also evaluate the proposed method (GAT with ResNet-$18$ architecture) on the DDN attack \cite{rony2019decoupling} for CIFAR-$10$ dataset, and obtain a mean $\ell_2$ norm of $0.805$ for adversarial perturbations, compared to $0.762$ as obtained with the Carlini and Wagner (C\&W) $\ell_2$ attack, indicating that the latter is stronger. Thus, we primarily utilise the C\&W attack for the evaluation of defense models on $\ell_2$ norm-constrained adversaries.
\begin{table}
\caption{Prediction accuracy (\%) of our model (GAT) in various targeted and untargeted White-Box attack settings. Among the targeted attacks, we consider 1000-step Least Likely attack and 1000-step Random target attack. In the second partition, worst case robustness against multiple random restarts is reported. We consider a 1000-sample subset of CIFAR-10 and ImageNet-100 datasets for the random restarts experiments.}
\label{table:targeted}
\setlength\tabcolsep{3pt}
\resizebox{1.0\linewidth}{!}{
\begin{tabular}{@{}l|cc|cc|cc@{}}
\toprule
               \multicolumn{1}{l|}{\textbf{Attack}} & \multicolumn{2}{c|}{\textbf{CIFAR-10}} & \multicolumn{2}{|c|}{\textbf{ImageNet-100}} & \multicolumn{2}{c}{\textbf{MNIST}} \\ 
                & $500$-step      & $1000$-step      & $500$-step      & $1000$-step     & $500$-step     & $1000$-step     \\ \midrule
PGD-Targeted (Least Likely class)        & 79.50             & 79.50              & 66.12             & 66.02              & 99.03             & 99.03              \\
PGD-Targeted (Random class)      & 74.56             & 74.37              & 63.80             & 64.10              & 98.86             & 98.84              \\
PGD-Untargeted       & 53.04             & 53.04              & 37.52             & 37.52              & 94.37             & 94.37 \\
\midrule
                & $1$-RR      & $1000$-RR      & $1$-RR      & $500$-RR      & $1$-RR      & $1000$-RR      \\ \midrule
PGD 50-step, r-RR & 53.20 & 52.10 & 38.70 & 38.30 & 95.46 & 92.20 \\
\bottomrule
\end{tabular}}
\end{table}

We evaluate our proposed approach on $1000$-step PGD targeted and untargeted attacks. The results of these experiments are presented in Table-\ref{table:targeted}. Targeted attacks are weaker than untargeted attacks, thereby resulting in a higher accuracy. We note that the attack converges within $1000$-steps based on the observation that drop in accuracy between $500$-step attack and $1000$-step attack is marginal. 

We present the worst-case accuracy of GAT-trained models across multiple random restarts of PGD 50-step attack in the second partition of Table-\ref{table:targeted}. The results are shown on a 1000-image subset of CIFAR-10 and ImageNet-100 test sets and on the full MNIST test set.  In order to find the worst-case accuracy, we continue restarts until accuracy stabilizes. We note that the robustness of the proposed defense is not broken by attacks using random restarts, thereby demonstrating the absence of gradient masking.

\subsection{Performance against Black-Box and gradient-free attacks}
\begin{table}
\caption{Prediction accuracy (\%) of different defenses in \textbf{FGSM Black-Box attack} setting. Source model for attack is specified in the column headings. Target model is specified in each row.}
\label{table:bb_fgsm_all}
\setlength\tabcolsep{3pt}
\resizebox{1.0\linewidth}{!}{
\begin{tabular}{@{}l|ccc|ccc|ccc@{}}
\toprule
                \textbf{Method}& \multicolumn{3}{c|}{\textbf{CIFAR-10}} & \multicolumn{3}{|c|}{\textbf{ImageNet-100}} & \multicolumn{3}{c}{\textbf{MNIST}} \\ 
                & Clean & VGG11      & ResNet18     & Clean & AlexNet      & ResNet18   & Clean & BB-MNIST      & M-LeNet      \\ \midrule
Normal     & 92.30  & 37.09 & 15.98 & 81.44                      & 63.82 & 8.22  & 99.05 & 38.62 & 16.58 \\
RFGSM-AT \cite{tramer2017ensemble}  & 89.66 & 82.62 & 85.74 & 78.46                      & 74.82 & 74.18 & 99.37 & 93.79 & 92.44 \\
FBF \cite{wong2020fast}       & 82.83 & 78.19 & 80.41 & 57.32                      & 56.12 & 56.22 & 99.30  & 95.56 & 95.19 \\
R-MGM \cite{baburaj2019regularizer}     & 82.29 & 78.18 & 79.99 & 64.84                      & 63.26 & 63.60  & 99.04 & 95.52 & 95.19 \\
GAT (\textbf{Ours}) & 80.49 & 76.95 & 78.54 & 67.98 & 65.94 & 65.98 & 99.37 & 96.51 & 96.49 \\
\midrule
PGD-AT \cite{madry-iclr-2018}    & 82.67 & 78.91 & 80.53 & 68.62                      & 67.02 & 67.34 & 99.27 & 95.68 & 96.27 \\
TRADES \cite{zhang2019theoretically}   & 81.73 & 78.28 & 79.65 & 62.88                      & 61.42 & 61.42 & 99.32 & 96.49 & 96.08 \\

 \bottomrule
\end{tabular}
}
\end{table}
\begin{table}
\caption{Prediction accuracy (\%) of different models in \textbf{PGD-7 step Black-Box attack} setting. Source model for attack is specified in the column headings. Target model is specified in each row.}
\label{table:bb_pgd_all}
\setlength\tabcolsep{3pt}
\resizebox{1.0\linewidth}{!}{
\begin{tabular}{@{}l|ccc|ccc|ccc@{}}
\toprule
                \textbf{Method}& \multicolumn{3}{c|}{\textbf{CIFAR-10}} & \multicolumn{3}{|c|}{\textbf{ImageNet-100}} & \multicolumn{3}{c}{\textbf{MNIST}} \\ 
                & Clean & VGG11      & ResNet18     & Clean & AlexNet      & ResNet18   & Clean & BB-MNIST      & M-LeNet      \\ \midrule

Normal     & 92.30  & 20.88 & 0.00     & 81.44                      & 72.48 & 0.04  & 99.05 & 8.03  & 0.01  \\
RFGSM-AT \cite{tramer2017ensemble}  & 89.66 & 84.96 & 87.36 & 78.46                      & 76.58 & 76.24 & 99.37 & 94.93 & 89.47 \\
FBF \cite{wong2020fast}       & 82.83 & 79.62 & 81.32 & 57.32                      & 56.66 & 56.84 & 99.30  & 96.53 & 96.51 \\
R-MGM \cite{baburaj2019regularizer}     & 82.29 & 79.33 & 80.92 & 64.84                      & 63.88 & 64.20  & 99.04 & 96.03 & 96.28 \\

GAT (\textbf{Ours}) & 80.49 & 77.88 & 79.25 & 67.98 & 66.58 & 66.80  & 99.37 & 97.25 & 97.52\\
\midrule
PGD-AT \cite{madry-iclr-2018}    & 82.67 & 79.68 & 81.26 & 68.62                      & 67.66 & 67.90  & 99.27 & 96.46 & 94.91 \\
TRADES \cite{zhang2019theoretically}    & 81.73 & 79.04 & 80.22 & 62.88                      & 62.28 & 62.26 & 99.32 & 96.98 & 94.86 \\
\bottomrule
\end{tabular}
}
\end{table}
\begin{table}
\caption{Prediction accuracy (\%) of different defenses against the query based Black Box attack, Square.}
\label{table:square}
\setlength\tabcolsep{3pt}
\resizebox{1.0\linewidth}{!}{
\begin{tabular}{l|cc|cc|cc}
\toprule
\multicolumn{1}{l}{\textbf{Method}}      & \multicolumn{2}{|c}{\textbf{~~CIFAR-10~~}} & \multicolumn{2}{|c}{\textbf{~~ImageNet-100~~}}    & \multicolumn{2}{|c}{\textbf{~~MNIST~~}} \\
\multicolumn{1}{l}{}      & \multicolumn{1}{|l}{~~~~Clean~~~~}    & ~~~~Square~~~~   & \multicolumn{1}{|l}{~~~~Clean~~~~}             & ~~~~Square~~~~ & \multicolumn{1}{|l}{~~~~Clean~~~~}  & ~~~~Square~~~~  \\
\midrule
Normal                    & 92.30              & 0.16              & 81.44                      & 0.66            & 99.05           & 0.02             \\
RFGSM-AT \cite{tramer2017ensemble}~~~~~~~~                 & 89.66             & 43.01             & 78.44                      & 41.60            & 99.37           & 84.13            \\
FBF \cite{wong2020fast}   & 82.83             & 52.46             & 57.32                      & 28.38           & 99.30            & 79.66            \\
R-MGM \cite{baburaj2019regularizer} & 82.29             & 52.87             & 64.82                      & 38.34           & 99.04           & 88.89            \\
GAT (\textbf{Ours})                & 80.49             & 53.62             & 67.98 & 39.06           & 99.37           & 91.03 \\
\midrule
PGD-AT \cite{madry-iclr-2018}                   & 82.67             & 52.64             & 68.60                       & 45.50            & 99.27           & 91.99            \\
TRADES \cite{zhang2019theoretically}                   & 81.73             & 54.87             & 62.86                      & 40.30            & 99.32           & 92.60             \\

\bottomrule
\end{tabular}
}
\end{table}
The results on FGSM and PGD $7$-step Black-box attacks are presented in Table-\ref{table:bb_fgsm_all} and Table-\ref{table:bb_pgd_all} respectively. We consider two sources for Black-box attacks; the first with the same architecture as the source model, and the second with a different architecture. Across all three datasets, accuracy on black-box attacks closely tracks the accuracy on clean samples, and is higher than that of white-box attacks. This confirms that there is no issue of gradient masking in the proposed model, which could potentially generate weaker adversaries, thereby creating a false sense of robustness. 

We present results on the query-based black-box attack, Square \cite{andriushchenko2019square} in Table-\ref{table:square}. We note that across most defenses, this attack is weaker than PGD-500 step attack for CIFAR-$10$ and ImageNet-$100$, whereas it is stronger than the same for MNIST. The proposed defense achieves improved robustness compared to existing single-step adversarial defenses on MNIST and CIFAR-$10$ datasets. For Imagenet-$100$, although R-FGSM achieves better accuracy, it is significantly more susceptible to white-box attacks presented in Table-\ref{table:imagenet_whitebox}. The performance of the proposed single-step defense GAT against the strong query based Square attack shows that the robustness of the proposed defense is not a result of gradient masking. 

Further, we evaluate the proposed defense against SPSA \cite{uesato2018adversarial}, a gradient-free attack that utilises a numerical approximation of gradients by sampling function values along random directions. We use the following standard hyperparameters for attack generation using SPSA: learning rate $= 0.01$, $\delta=0.01$, number of iterations $=100$ and number of samples to approximate the average gradient $=128$. On the CIFAR-$10$ dataset, the proposed single-step GAT defense with ResNet-$18$ architecture achieves $56.59\%$ accuracy against SPSA, compared to $53.62\%$ on the Square attack. Since the Square attack is stronger than SPSA, we use the former to present gradient-free attack evaluation of all defense methods in Table-\ref{table:square}. 

\subsection{Performance Against Adaptive Adversaries}

\begin{table}
\caption{Prediction accuracy (\%) of our proposed single-step defense GAT against adaptive attacks constructed using diverse settings (rows).}
\label{table:adaptive}
\setlength\tabcolsep{3pt}
\resizebox{1.0\linewidth}{!}{
\begin{tabular}{c|c|cc|cc|cc}
\toprule
\multicolumn{1}{c|}{$\lambda_{attack}$}       & \multicolumn{1}{c}{Decay}                 & \multicolumn{2}{|c}{\textbf{CIFAR-10}}                                       & \multicolumn{2}{|c}{\textbf{ImageNet-100}}                                       & \multicolumn{2}{|c}{\textbf{MNIST}}                                          \\
\multicolumn{1}{c}{} & \multicolumn{1}{|c}{Iterations} & \multicolumn{1}{|c}{~~~GA-CE~~~} & \multicolumn{1}{c}{~~~GAMA~~~} & \multicolumn{1}{|c}{~~~GA-CE~~~} & \multicolumn{1}{c}{~~~GAMA~~~} & \multicolumn{1}{|c}{~~~GA-CE~~~} & \multicolumn{1}{c}{~~~GAMA~~~} \\
\midrule
0                                   & no decay                                      & 52.93                                & 48.40                                & 36.66                                & 30.14                                & 93.22                                & 93.06                                \\
$\lambda_{defense,init}$                         & no decay                                      & 52.83                                & 49.78                                & 35.76                                & 32.80                                & 93.07                                & 93.20                                \\
$\lambda_{defense,final}$                        & no decay                                      & 53.63                                & 51.74                                & 35.64                                & 35.18                                & 93.09                                & 93.15                                \\
50                                  & 25                                            & 52.80                                & \textbf{47.76}                                & 36.58                                & \textbf{29.24}                                & \textbf{93.02}                                & \textbf{93.00}                                \\
100                                 & 25                                            & 52.75                                & 47.92                                & 36.66                                & 29.34                                & 93.13                                & 93.05                                \\
50                                  & 50                                            & \textbf{52.73}                                & 47.88                                & 36.56                                & 29.40                                & 93.20                                & 93.02                                \\
100                                 & 50                                            & 52.84                                & 47.97                                & 36.58                                & 29.30                                & 93.23                                & 93.11                                \\
50                                  & 100                                           & 53.00                                & 50.64                                & 36.58                                & 32.22                                & 93.19                                & 93.15                                \\
100                                 & 100                                           & 53.34                                & 51.68                                & \textbf{35.52}                                & 33.54                                & 93.39                                & 93.35                                \\

\bottomrule
\end{tabular}}
\end{table}
\begin{figure}[t]
    \centering
    \includegraphics[width=1.0\linewidth]{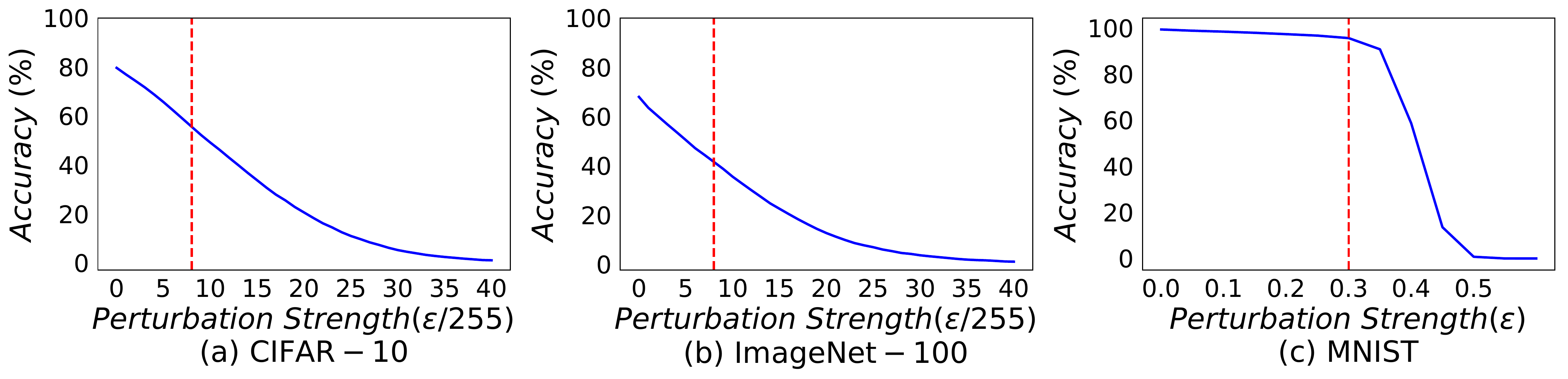}
    \vspace{-0.3cm}
    \caption{Plot showing the variation of accuracy on $7$-step PGD samples across the magnitude of perturbation, $\varepsilon$ for the proposed single-step defense, GAT. Accuracy goes to zero for large $\varepsilon$ across all datasets, indicating the absence of gradient masking.}
    \label{fig:accvseps}
\end{figure}

\begin{figure}[t]
    \centering
    \includegraphics[width=1.0\linewidth]{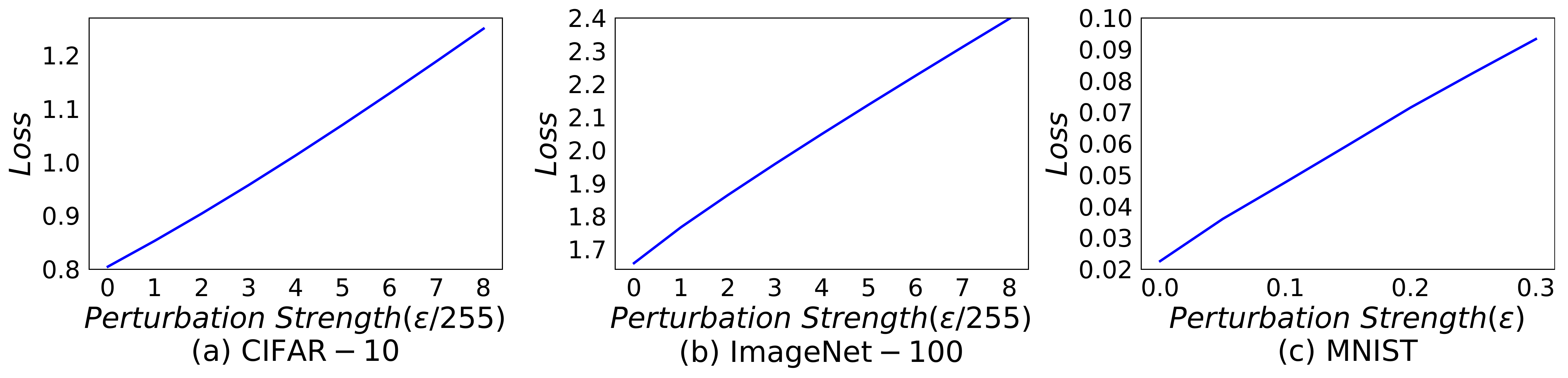}
    \vspace{-0.3cm}
    \caption{Plot showing the variation of loss on FGSM samples across the magnitude of perturbation, $\varepsilon$ for the proposed single-step defense, GAT. Loss increases monotonically with an increase in $\varepsilon$ across all datasets, indicating the absence of gradient masking.}
    \label{fig:lossvseps}
\end{figure}
Since we consider the framework of worst-case adversarial robustness, we assume that the adversary has complete knowledge of the defense mechanism employed. Thus, it is crucial to evaluate our model on adaptive adversaries as well \cite{carlini2019evaluating}. We consider strong multi-step PGD attacks, where the adversary attempts to maximise the proposed loss at each step. More precisely, the adversaries are generated using the proposed GA-CE or GAMA loss, where the first term is either cross-entropy loss or maximum margin loss, and the second term is the  squared $\ell_2$ distance between the softmax outputs of the original clean image and the adversarial sample generated in the previous step. We evaluate the proposed defense against diverse settings of the attack, using the cross-entropy loss in one case (GA-CE) and maximum-margin loss in the second (GAMA). We present our experiments and results in Table-\ref{table:adaptive}. We consider a case without $\lambda$ decay over iterations, where the $\lambda$ used in the attack ($\lambda_{attack}$) is same as the $\lambda$ used in defense ($\lambda_{defense,init}$). We also consider a case where $\lambda_{attack} = \lambda_{defense,final} $ which is the same as $\lambda_{defense,init}$ multiplied by the step-up factor. In general, the cross-entropy based attack is weaker than the margin based attack. The adaptive attacks are significantly stronger than standard PGD-based attacks. However, they are not as strong as the AutoAttack (AA). We note that it is not fair to compare the strength of a single attack with an ensemble of attacks, which declares a given data sample to be robust only if it passes all the attacks in the ensemble. Therefore, although we generate strong adaptive attacks, robustness of the proposed approach does not deteriorate further compared to AutoAttack. 

\subsection{Sanity checks to ensure absence of gradient masking}

We observe from the above experiments that iterative attacks are stronger than single-step attacks (Table-\ref{table:cifar10_whitebox} in the main paper, Tables-\ref{table:imagenet_whitebox} and \ref{table:mnist_whitebox}). Furthermore, white-box attacks are stronger than black-box attacks (Table-\ref{table:cifar10_whitebox} in the main paper, Tables-\ref{table:imagenet_whitebox}, \ref{table:mnist_whitebox}, \ref{table:bb_fgsm_all} and \ref{table:bb_pgd_all}). In the plot shown in Fig.\ref{fig:accvseps}, we increase the value of $\varepsilon$ and observe that unbounded attacks are able to reach 100\% attack success rate for all datasets. Also, as shown in Fig.\ref{fig:lossvseps}, the loss monotonically increases with an increase in the perturbation size of the FGSM attack. These tests confirm that the model is truly robust, and the observed robustness is not a result of gradient masking. 

\end{document}